\RequirePackage{ifthen}
\newboolean{IJOO}
\setboolean{IJOO}{false}

\newif\ifIJOOincludeMain
\newif\ifIJOOincludeAppendix
\newif\ifIJOOmainonly
\newif\ifIJOOappendixonly
\IJOOincludeMaintrue
\IJOOincludeAppendixtrue
\IJOOmainonlyfalse
\IJOOappendixonlyfalse
\providecommand{\IJOOMainLabelsFile}{main_body}
\providecommand{\IJOOAppendixLabelsFile}{appendix_only}
\ifdefined\IJOOMainOnly
  \IJOOincludeMaintrue
  \IJOOincludeAppendixfalse
  \IJOOmainonlytrue
\fi
\ifdefined\IJOOAppendixOnly
  \IJOOincludeMainfalse
  \IJOOincludeAppendixtrue
  \IJOOappendixonlytrue
\fi

\ifthenelse{\boolean{IJOO}}
{
\documentclass[ijoo,sglanonrev]{informs4}
\RequirePackage{tgtermes}
\RequirePackage{newtxtext}
\RequirePackage{newtxmath}
\RequirePackage{bm}
\RequirePackage{endnotes}
\usepackage{eqndefns-left} 
\OneAndAHalfSpacedXII 

\EquationsNumberedThrough    

\TheoremsNumberedThrough     
\ECRepeatTheorems  %

\MANUSCRIPTNO{IJOO-0001-2024.00}

\usepackage{hyperref}
\newenvironment{prf}[1][]
{\proof{Proof. }}
{\hfill\ensuremath{\Halmos} \endproof}

\newenvironment{prfc}[1][]
{\proof{Proof #1.}}
{\hfill\ensuremath{\Halmos} \endproof}

} 
{
\documentclass[11pt]{article}
\usepackage[margin=1in]{geometry}

\usepackage{hyperref}
\usepackage{amsthm}

\newtheorem{theorem}{Theorem}

\newtheorem{assumption}{Assumption}
\newtheorem*{remark}{Remark}
\newtheorem{proposition}[theorem]{Proposition}

\newtheorem{lemma}[theorem]{Lemma}
\newtheorem*{example}{Example}

\newenvironment{prf}[1][]
{\begin{proof}}
{\end{proof}}

\newenvironment{prfc}[1][]
{\begin{proof}[Proof #1]}
{\end{proof}}

}

\ifIJOOmainonly
\usepackage{xr-hyper}
\IfFileExists{\IJOOAppendixLabelsFile.aux}{%
  \externaldocument{\IJOOAppendixLabelsFile}%
}{%
  \typeout{IJOO main-only mode: compile \IJOOAppendixLabelsFile.tex once if you want references to the appendix to resolve.}%
}
\fi
\ifIJOOappendixonly
\usepackage{xr-hyper}
\IfFileExists{\IJOOMainLabelsFile.aux}{%
  \externaldocument{\IJOOMainLabelsFile}%
}{%
  \typeout{IJOO appendix-only mode: compile \IJOOMainLabelsFile.tex once first to import labels from the main paper.}%
}
\fi

\usepackage{thm-restate}

\usepackage{amsmath}
\usepackage{amssymb}

\usepackage{algorithm}
\usepackage[noend]{algorithmic}
\usepackage{enumitem}
\usepackage{tikz}
\usepackage{subcaption}
\usepackage{makecell}
\usepackage{booktabs}       
\usepackage{pdflscape}   
\usepackage{adjustbox}   
\usepackage{verbatim}
\usepackage{float}

\graphicspath{{../}{./}}
\makeatletter
\def\input@path{{../}{./}}%
\makeatother

\usepackage[capitalize,noabbrev]{cleveref}
\newtheorem{model}{Model}
\newtheorem{fact}{Fact}

\usepackage{mathtools}
\DeclareMathOperator{\tr}{tr}
\DeclareMathOperator{\rk}{rank}
\DeclareMathOperator{\sp1}{Span}

\DeclareMathOperator{\sign}{sign}
\DeclareMathOperator{\supp}{Supp}

\DeclareMathOperator{\mo}{\mathcal{O}}

\newcommand{\R}{\mathbb{R}}

\newcommand{\abs}[1]{\left\lvert#1\right\rvert}
\newcommand{\norm}[1]{\left\lVert#1\right\rVert_2}
\newcommand{\znorm}[1]{\left\lVert#1\right\rVert_0}
\newcommand{\onorm}[1]{\left\lVert#1\right\rVert_1}
\newcommand{\fnorm}[1]{\left\lVert#1\right\rVert_F}
\newcommand{\maxnorm}[1]{\left\lVert#1\right\rVert_{\infty}}
\newcommand{\infnorm}[1]{\left\lVert#1\right\rVert_{\infty}}
\newcommand{\ottnorm}[1]{\left\lVert#1\right\rVert_{1\rightarrow2}}
\allowdisplaybreaks

\def\mp{{\mathbb P}}
\def\me{{\mathbb E}}

\def\sg{{\mathcal{SG}}}

\newcommand{\ttl}{A Randomized Algorithm for Sparse PCA\\ based on the Basic SDP Relaxation}

\newcommand{\bstrct}{Sparse Principal Component Analysis (SPCA) is a fundamental technique for dimensionality reduction, and is NP-hard. 
In this paper, we introduce a randomized approximation algorithm for SPCA, which is based on the basic SDP relaxation.
Our algorithm takes an (approximate) SDP solution, constructs one deterministic sparse solution and several randomized solutions, and outputs the best among them.
Our algorithm has an approximation ratio of at most the sparsity constant with high probability, if called enough times. 
Under a technical assumption, which is consistently satisfied in our numerical tests, the average approximation ratio is also bounded by $\mathcal{O}(\log{d})$, where $d$ is the number of features.
We show that this technical assumption is satisfied if the SDP solution is low-rank, or has exponentially decaying eigenvalues.
We then present two classes of instances for which this technical assumption holds. 
We also demonstrate that in a covariance model, which generalizes the spiked Wishart model, the deterministic solution in our algorithm achieves a near-optimal approximation ratio.
We demonstrate the efficacy of our algorithm through numerical tests on real-world datasets.
}

\newcommand{\kywrds}{Sparse PCA, Randomized algorithm, Semidefinite Programming}

\ifthenelse{\boolean{IJOO}}
{
\usepackage[sort&compress]{natbib}
 \bibpunct[, ]{[}{]}{,}{n}{}{,}%
\TITLE{\ttl}
\RUNTITLE{A Randomized Algorithm for Sparse PCA}
\ARTICLEAUTHORS{%
\AUTHOR{Anonymous Author(s)}%
\AFF{Anonymous Institution(s)}%
}%
\RUNAUTHOR{Anonymous}

\ABSTRACT{\bstrct}
}
{

\DeclareMathOperator*{\argmax}{arg\,max}

\title{\ttl}

\author{Alberto Del Pia
\thanks{Department of Industrial and Systems Engineering \& Wisconsin Institute for Discovery,
University of Wisconsin-Madison.
E-mail: {\tt delpia@wisc.edu}}
\and
Dekun Zhou
\thanks{Department of Industrial and Systems Engineering \& Wisconsin Institute for Discovery, University of Wisconsin-Madison.
E-mail: {\tt dzhou44@wisc.edu}}}
}

\newcommand{\IJOOInputProofAppendix}{%
\ifthenelse{\boolean{IJOO}}{%
\begin{APPENDICES}
\section{Proofs in \cref{sec:ra for spca}}
	\label{sec:proof of ra}
	In this section, we present the proofs that we owe in \cref{sec:ra for spca}.

    \subsection{Proofs in \cref{sec:app_guarantee}}
    \label{sec:proof of app guarantee}
    In this section, we prove \cref{thm:spca rand alg v2,thm:spca rand alg}.
    To do this, we need to introduce several new notation and random events.

	Firstly, we show that \cref{alg:randomized spca simple} produces a $k$-sparse vector with a high probability. 
	\begin{proposition}
		\label{prop:sum of epsilon}
		Let $S$ be the set in \cref{alg:randomized spca simple}.
		Define a random event 
        {\setlength{\abovedisplayskip}{2pt}
        \setlength{\belowdisplayskip}{2pt}%
		\begin{equation}
			\label{def:def of A}
			\mathcal{A}:=\left\{\abs{S} \le k\right\}.
		\end{equation}}
		Then, $\mathcal{A}$ holds true with probability at least $1 - \exp\{- c k\}$ for some absolute constant $c > 0$.
	\end{proposition}

    The proof of this property relies on the well-known multiplicative Chernoff bound:
	\begin{lemma}[\cite{mitzenmacher2017probability}, Theorem~4.4 and the remark after Corollary~4.6]
		\label{prop:multiplicative chernoff}
		Let $X = \sum_{i = 1}^d X_i$, where $X_i$'s are independent Bernoulli (i.e., 0-1) random variables. 
		Let $L_u$ be a number such that $L_u \ge \me X$. 
		Then, for any $\delta > 0$, we have
        {\setlength{\abovedisplayskip}{0pt}
  \setlength{\belowdisplayskip}{0pt}%
		\begin{equation*}
		    \mp\left(X \ge (1+\delta) L_u\right) \le \left(\frac{e^\delta}{(1+\delta)^{1+\delta}}\right)^{L_u}.
		\end{equation*}}
	\end{lemma}
	
	\begin{prfc}[of \cref{prop:sum of epsilon}]
		We use \cref{prop:multiplicative chernoff} to prove this statement.
		It is equivalent to showing that $\sum_{i = 1}^d \epsilon_i \le k$ with probability at least $1 - \exp\{- c k\}$ for some $c>0$, where $\epsilon_i$'s are the Bernoulli random variables in \cref{alg:randomized spca simple}.
		Define $L_u := 3/4\cdot k$, it is clear that $\me \sum_{i=1}^d \epsilon = \sum_{i=1}^d p_i \le L_u$.
		By \cref{prop:multiplicative chernoff}, we pick $\delta:=1/3$, and obtain the desired result with $c = -1/4 - \log(3/4) > 0.037$. 
     \end{prfc}

     Now, we are ready to prove \cref{thm:spca rand alg v2}.
     \begin{prfc}[of \cref{thm:spca rand alg v2}]
	Define $i^*:=\argmax_{i\in [d]} A_{ii}$, and it suffices to show that $z^\top A z \ge A_{i^*i^*} = \infnorm{A}$ with high probability.
	Indeed, by H\"older's inequality, for any feasible solution $W$ to \ref{prob SDP SPCA}, we have $\tr(AW)\le \infnorm{A} \cdot \onorm{W} \le k\infnorm{A} = kA_{i^*i^*}$, where the last equality follows from the fact that for any $i\ne j$, one have $A_{ij}^2 \le A_{ii} A_{jj} \le \max\{A_{ii}, A_{jj}\}^2$ due to $A\succeq 0$.
	
	For each $i\in [N]$, let $E_i$ be the event that $\epsilon_{i^*}=1$ and $\sum_{j\ne i^*} \epsilon_j \le k-1$ in the $i$-th call of \cref{alg:randomized spca simple}. Since $A\succeq 0$, line~8 of \cref{alg:randomized spca simple} only augments the sampled support when its size is smaller than $k$. Hence, on $E_i$, the resulting feasible support contains $i^*$, and the corresponding output $z_i$ satisfies $z_i^\top A z_i \ge A_{i^*i^*}$.
	Moreover,
    \begin{equation*}
        \mp(E_i)=p_{i^*} \cdot \mp\left(\sum_{j\ne i^*} \epsilon_j \le k-1\right),
    \end{equation*}
	because $\epsilon_{i^*}$ is independent of $\{\epsilon_j\}_{j\ne i^*}$. As in the proof of \cref{prop:sum of epsilon},
    \begin{equation*}
        \me \sum_{j\ne i^*} \epsilon_j = \sum_{j\ne i^*} p_j \le \sum_{j=1}^d p_j \le \frac{3k}{4}.
    \end{equation*}
	Therefore, by Markov's inequality,
    \begin{equation*}
        \mp\left(\sum_{j\ne i^*} \epsilon_j \ge k\right)
        \le \frac{\me \sum_{j\ne i^*} \epsilon_j}{k}
        \le \frac{3}{4},
    \end{equation*}
	and thus $\mp(E_i)\ge p_{i^*}/4$. The events $E_1,\ldots,E_N$ are independent across the $N$ calls of \cref{alg:randomized spca simple}, so
    \begin{equation*}
        \mp\left(\bigcup_{i=1}^N E_i\right)
        = 1 - \prod_{i=1}^N \left(1-\mp(E_i)\right)
        \ge 1 - \left(1-\frac{p_{i^*}}{4}\right)^N
        \ge 1 - \exp\left\{-\frac{Np_{i^*}}{4}\right\}.
    \end{equation*}
	Finally, by the definition of $p_{i^*}$ and the fact that $dA_{i^*i^*}\ge \tr(A)$,
    \begin{equation*}
        p_{i^*} \ge \frac{kA_{i^*i^*}}{12\tr(A)} \ge \frac{k}{12d},
    \end{equation*}
	which yields
    \begin{equation*}
        \mp\left(\bigcup_{i=1}^N E_i\right) \ge 1 - \exp\left\{-\frac{kN}{48d}\right\}.
    \end{equation*}
	Whenever $\bigcup_{i=1}^N E_i$ occurs, one of the feasible candidates considered by \cref{alg:multi_ra} has objective value at least $A_{i^*i^*}$. Since \cref{alg:multi_ra} returns the best feasible candidate, the returned vector $z$ satisfies $z^\top A z \ge A_{i^*i^*}$.
\end{prfc}

	To prove \cref{thm:spca rand alg}, we need an additional useful probabilistic property of the uniform random vector $g$ that we will be used in the proof. 
	\begin{proposition}
		\label{prop:alg rv u}
		Let $g\in \R^d$ be a random vector such that $g_i \stackrel{\text{i.i.d.}}{\sim} \text{Uniform}(-\sqrt{3}, \sqrt{3})$.
		Assume a matrix $W\succeq 0$, and denote $U:=\sqrt{W}$, and let $U = (u_1, u_2, \ldots, u_d)$.
		Define the random event as follows:
		\begin{equation}
			\label{def:def of random event B}
			\mathcal{B}:=\left\{\max_{i\in [d], \norm{u_i} > 0} \frac{|u_i^\top g|^2}{\norm{u_i}^2} \le C\log{d}  \right\}
		\end{equation}
		Then, $\mathcal{B}$ holds true for some absolute constant $C > 0$, with probability at least $1 - 2d^{-3}$.
	\end{proposition}

	\begin{prfc}[of \cref{prop:alg rv u}]
		WLOG we assume that $u_i \ne 0_d$ for every $i\in [d]$.
		Since each $g_i$ is a bounded random variable, we see that $g_i \sim \sg(3)$, i.e., $\me \exp\{t g_i\}\le \exp\{3t^2 / 2\}$.
		Moreover, as each $g_i$ has i.i.d.~entries, it is clear that $u_i^\top g \sim \sg (3 \norm{u_i}^2)$, and hence by standard Chernoff bound (see, e.g., Section~4.2 in \cite{mitzenmacher2017probability}), for an arbitrary $s > 0$, $\mp\left( |u_i^\top g| > s \right) \le 2 \exp\left\{ - s^2 / (6 \norm{u_i}^2) \right\}$.
		By union bound, for any $t > 0$, we obtain that
		\begin{equation*}
            \mp\left( \max_{i \in [d]} \frac{|u_i^\top g|}{\norm{u_i}} > t \right)
			\le \sum_{i = 1}^d \mp\left(  \frac{\abs{u_i^\top g}}{\norm{u_i}} > t\right)\le \sum_{i = 1}^d 2 \exp\left\{ - \frac{t^2}{6 } \right\}
			= 2d \exp\left\{ - \frac{t^2}{6} \right\}.
		\end{equation*}
		Setting $t:=2\sqrt{6}  \sqrt{\log{d}}$ concludes the proof.     
    \end{prfc}

    Now, we are ready to give a formal statement of \cref{thm:spca rand alg}:

    	\begin{theorem}
		\label{thm:spca rand alg formal}
		Let the notation be the same as in \cref{thm:spca rand alg}, but only assuming that $A\in \R^{d\times d}$ is symmetric with $\norm{A} = 1$.
		Denote random events $\mathcal{A}$ and $\mathcal{B}$ as defined in \eqref{def:def of A} and \eqref{def:def of random event B}, respectively, and denote $c>0$ and $C>0$ the absolute constants in \cref{prop:sum of epsilon} and \cref{prop:alg rv u}, respectively.
		\begin{enumerate}[leftmargin = *]
			\item Suppose that the input matrix $A$ has non-negative diagonal entries, then one has
			\begin{equation*}
			        C \log{d} \left[1 + \frac{9\textup{SSR}^2}{4k}\right]   \me \left[ z^\top A z \vert \mathcal{A}\cap \mathcal{B} \right]
				 \ge \left[1 - \mathcal{O}\left(\frac{1}{d}\right)\right](x^*)^\top A x^* - \epsilon - e^{-ck + 2\log{(\frac{2d}{k})}}.
			\end{equation*}
			\item Suppose, in addition, that the input matrix $A$ is positive semidefinite, and that $ck\ge 3\log{(d/k)} + \log\log{d}$. 
			Then, 
			\begin{equation*}
			    \begin{aligned}
			        C \log{d}  \left[1 + \frac{9\textup{SSR}^2}{4k}\right]  \me \left[ z^\top A z \vert \mathcal{A}\cap \mathcal{B} \right]
				 \ge \left[1  -  \mathcal{O}\left(\frac{1}{\log{d}} \right)\right] (x^*)^\top A x^* - \epsilon.
			    \end{aligned}
			\end{equation*}
		\end{enumerate}
	\end{theorem}

  It is clear that \cref{thm:spca rand alg formal} indeed implies \cref{thm:spca rand alg}, by setting $\mathcal{R}:= \mathcal{A} \cap \mathcal{B}$, and it is clear that the probability that $\mathcal{R}$ occurs is lower bounded by $ 1 - \exp\{-ck\} - 2d^{-3}$ by \cref{prop:sum of epsilon,prop:alg rv u}.
  Note that, the lower bound on $ck$ in part~2 is used only to simplify the explicit residual term from part~1 to $\mathcal{O}(1/\log d)$; any stronger lower bound on $ck$ would yield a correspondingly smaller lower-order term.
	
	We are now ready to prove \cref{thm:spca rand alg formal}.
	
	\begin{prfc}[of \cref{thm:spca rand alg formal}]
        In the proof, denote by $\sqrt{W^*} = (u_1, u_2, \dots, u_d)$, so that $\textup{SSR} = \sum_{i = 1}^d \norm{u_i}$.
        We first point out a fact regarding \cref{alg:randomized spca simple}: After executing line 6, i.e., setting the set $S$ to be the set $\{i\in [d]: \epsilon_i = 1\}$, if one draws a random vector $g\in \R^d$ such that its entries are all i.i.d.~Uniform in $[-\sqrt{3}, \sqrt{3}]$, and defines a random vector $x\in\R^d$ such that $x_i = 0$ for $i\in S^c$, and $x_i = u_i^\top g / p_i$ for $i\in S$, then one can see that $z^\top A z \ge x^\top A x / \norm{x}^2$.
        Indeed, this is implied by the fact that $\supp(x) \subseteq \supp(z)$, and $z$ is chosen to the vector with larger \ref{prob SPCA} objective on line~9.

        The ideas behind the rest of the proof are as follows: The expected objective value $x^\top A x$ is greater or equal to $\tr(AW^*)$.
        This, coupled with the facts that $z^\top A z \ge x^\top A x / \norm{x}^2$, $\tr(AW^*)\ge (x^*)^\top A x^* - \epsilon$, and upper bound of $\tr(AW^*)$, lead to approximation guarantees.

        \textbf{Expected value of $x^\top A x$.} Assuming that $p_i > 0$ for all $i\in [d]$.
        By definition, 
        {\setlength{\abovedisplayskip}{2pt}
  \setlength{\belowdisplayskip}{2pt}
        \begin{equation*}
		    x^\top A x = \sum_{i, j = 1}^d a_{ij} x_{i} x_j
			= \sum_{i,j = 1}^d a_{ij} \cdot \left(\frac{u_i^\top g}{p_i} \cdot \epsilon_i\right) \cdot \left(\frac{u_j^\top g}{p_j} \cdot \epsilon_j\right).
		\end{equation*}
        }
		For $i, j \in [d]$ and $i\ne j$, it holds that $\me x_i x_j = \me (u_i^\top g) (u_j^\top g) = u_i^\top \left(\me g g^\top\right) u_j = u_i^\top I_d u_j = u_i^\top u_j$, and for $i = j\in [d]$, we have that $\me x_i^2 = \me (u_i^\top g)^2 / p_i = u_i^\top \left(\me g g^\top\right) u_i / p_i = \norm{u_i}^2 / p_i.$
		Since $U = \sqrt{W}$, we have that $U^2 = U^\top U = W$, and hence $W_{ij} = u_i^\top u_j$.
		Combining all above facts, we obtain that
        {\setlength{\abovedisplayskip}{2pt}
  \setlength{\belowdisplayskip}{2pt}
		\begin{equation*}
		    \me x^\top A x 
			= \sum_{i = 1}^d \frac{a_{ii}}{p_i} \norm{u_i}^2 + \sum_{1\le i \ne j \le d} a_{ij} u_i^\top u_j
			= \tr\left(A W\right) + \sum_{i = 1}^d\left( \frac{1}{p_i} -  1  \right) a_{ii} \norm{u_i}^2 \ge \tr(AW^*).
		\end{equation*}
        }
        In the case where there exist some $p_i = \min\{1, 2/3 \cdot k \norm{u_i} / \textup{SSR} + 1/12\cdot k A_{ii} / \tr(A)\}= 0$, it is clear that $x_i = 0$ and $W_{ii} = \norm{u_i}^2 = 0$. 
		$W\succeq 0$ implies that $W_{ij} = W_{ji} = 0$ for all $j\in [d]$, and thus $\me x_i x_j = 0 = W_{ij}$, and thus $\me x^\top A x \ge \tr(AW^*)$ still holds.

        \textbf{Upper bound of $\tr(AW^*)$.}
        By law of total expectation, we obtain that
        {\setlength{\abovedisplayskip}{0pt}
  \setlength{\belowdisplayskip}{0pt}
        \begin{align*}
			\me x^\top A x & = \mp(\mathcal{A}) \cdot \me \left[ x^\top A x \vert \mathcal{A} \right] + \mp(\mathcal{A}^c) \cdot \me \left[ x^\top A x \vert \mathcal{A}^c \right]\notag\\
			& \le \mp(\mathcal{A}) \cdot \me \left[ x^\top A x \vert \mathcal{A} \right] + \exp\{-ck\} \cdot \norm{A} \cdot \me \left[ x^\top x \vert \mathcal{A}^c \right] \notag\\
			& = \mp(\mathcal{A}) \cdot \me \left[ x^\top A x \vert \mathcal{A} \right] + \exp\{-ck\} \cdot \me \left[ x^\top x \vert \mathcal{A}^c \right]
		\end{align*} 
        }
		We first upper bound $\mp(\mathcal{A}) \cdot \me \left[ x^\top A x \vert \mathcal{A} \right]$.
        Again by law of total expectation, 
        {\setlength{\abovedisplayskip}{0pt}
  \setlength{\belowdisplayskip}{0pt}
        \begin{align}
			\label{eqn:cond expectation}
			\mp(\mathcal{A}) \cdot \me \left[ x^\top A x \vert \mathcal{A} \right]
			& = \mp\left(\mathcal{A}\cap \mathcal{B} \right) \cdot \me \left[ x^\top A x \vert \mathcal{A}\cap \mathcal{B} \right] + \mp\left(\mathcal{A}\cap \mathcal{B}^c \right) \cdot \me \left[ x^\top A x \vert \mathcal{A}\cap \mathcal{B}^c \right] 
		\end{align}
        }
    We upper bound $\norm{x}^2$ conditioned on $\mathcal{A} \cap \mathcal{B}$ first. 
		Assume that $\mathcal{A}$ occurs, define $S:=\supp(x)$, and let $T:=\{i\in S:p_i = 1\}$.
    Notice that $\sqrt{W_{ii}^*} = \norm{u_i}$ and $p_i \ge 2/3\cdot k \norm{u_i}/\textup{SSR}$, and thus
        {\setlength{\abovedisplayskip}{2pt}
  \setlength{\belowdisplayskip}{0pt}
		\begin{align*}
				\norm{x}^2 
				&= \sum_{i\in T} (u_i^\top g)^2 + \sum_{i\in S\backslash T} \frac{(u_i^\top g)^2}{p_i^2}
	            \le \sum_{i\in T} (u_i^\top g)^2 +  \frac{9}{4k^2} \textup{SSR}^2 \sum_{i\in S\backslash T} \frac{(u_i^\top g)^2}{\norm{u_i}^2}\\
			& \stackrel{\mathcal{B}}{\le} \sum_{i\in S} (C \log{d}) \cdot \norm{u_i}^2  +  \frac{9}{4k^2} \textup{SSR}^2 \cdot k \cdot C \log{d}
            \le  C \log{d} \left[1 + \frac{9}{4k} \textup{SSR}^2\right].
		\end{align*}
        }
	        Define $\bar x:= x / \norm{x}$.
		Hence, on the event $\mathcal{A}\cap\mathcal{B}$, the vector $\bar x$ is feasible to \ref{prob SPCA}, and by the definition of $z$ on line~9 of \cref{alg:randomized spca simple} one has $z^\top A z \ge \bar x^\top A \bar x$. 
    Combining this with the upper bound on $\norm{x}^2$ obtained above, we get the pointwise inequality assuming $\mathcal{A}\cap \mathcal{B}$ occurs:
        {\setlength{\abovedisplayskip}{2pt}
  \setlength{\belowdisplayskip}{0pt}
			\begin{align*}
				x^\top A x
				= \norm{x}^2 \bar x^\top A \bar x 
				\le C \log{d}  \Big(1 + \frac{9\textup{SSR}^2}{4k} \Big)  z^\top A z.
			\end{align*}
	        }
        Taking conditional expectations yields
        {\setlength{\abovedisplayskip}{2pt}
  \setlength{\belowdisplayskip}{0pt}
			\begin{align*}
				\me [ x^\top A x \vert \mathcal{A}\cap \mathcal{B} ]
	            \le C \log{d}  \Big(1 + \frac{9\textup{SSR}^2}{4k} \Big)  \me \left[ z^\top A z \vert \mathcal{A}\cap \mathcal{B} \right].
			\end{align*}
	        }
		Since $A$ has non-negative diagonal entries in part~1 and $A\succeq 0$ in part~2, the optimal value $(x^*)^\top A x^*$ is non-negative in both cases.
		We are then ready to upper bound the second term in \eqref{eqn:cond expectation}. 
    Assume for now that $\me [ x^\top A x \vert \mathcal{A}\cap \mathcal{B}^c ] \ge 0$.
		Later, we will tackle  $\me [ x^\top A x \vert \mathcal{A}\cap \mathcal{B}^c ] < 0$.
		By \cref{prop:alg rv u}, it is clear that $\mp(\mathcal{B}^c) \le  2d^{-3}$, thus
        {\setlength{\abovedisplayskip}{2pt}
  \setlength{\belowdisplayskip}{0pt}
		\begin{align*}
			\mp\left(\mathcal{A}\cap \mathcal{B}^c \right)  \me \left[ x^\top A x \vert \mathcal{A}\cap \mathcal{B}^c \right]
			 \le \frac{2}{d^3}  \me \left[ x^\top A x \vert \mathcal{A}\cap \mathcal{B}^c \right]
             \le \frac{2}{d^3}  \left[ (x^*)^\top A x^* \right]  \me \left[ \norm{x}^2 \vert \mathcal{A}\cap \mathcal{B}^c \right].
		\end{align*}
        }
		Then, we give an upper bound for $\me \left[ \norm{x}^2 \vert \mathcal{A}\cap \mathcal{B}^c \right]$.
		By writing down $\norm{x}^2$ explicitly, and by the fact that $(u_i^\top g)^2 \le \norm{u_i}^2 \norm{g}^2$, we observe that
		{\setlength{\abovedisplayskip}{0pt}
  \setlength{\belowdisplayskip}{0pt}
        \begin{align*}
				&\quad \me \left[ \norm{x}^2 \vert \mathcal{A}\cap \mathcal{B}^c \right]
	            \le \me \left[ \sum_{i\in T} (u_i^\top g)^2 +  \frac{9}{4k^2} \textup{SSR}^2 \cdot \sum_{i\in S\backslash T} \frac{(u_i^\top g)^2}{\norm{u_i}^2}  \Bigg\vert \mathcal{A}\cap \mathcal{B}^c \right]\\
			&\le  \me \left[ \sum_{i\in T} \norm{u_i}^2 \norm{g}^2 +  \frac{9}{4k^2} \textup{SSR}^2 \cdot \sum_{i\in S\backslash T} \frac{\norm{u_i}^2 \norm{g}^2}{\norm{u_i}^2} \Bigg\vert \mathcal{A}\cap \mathcal{B}^c \right]\\
			& \le  \me \left[ \norm{g}^2 +  \frac{9}{4k^2} \textup{SSR}^2 \cdot \sum_{i\in S\backslash T} \norm{g}^2 \Bigg\vert \mathcal{A}\cap \mathcal{B}^c \right]
            \le  3d  + \frac{27 d}{4k^2}\textup{SSR}^2,
		\end{align*}
        }
		where the last inequality follows from $g_i\sim \text{Uniform}(-\sqrt{3}, \sqrt{3})$ and hence $\norm{g}^2\le 3d$.
		By now, we have obtained that
    {\setlength{\abovedisplayskip}{0pt}
  \setlength{\belowdisplayskip}{0pt}
    \begin{equation*}
        \mp\left(\mathcal{A}\cap \mathcal{B}^c \right) \me [ x^\top A x \vert \mathcal{A}\cap \mathcal{B}^c ]
        \le (x^*)^\top A x^* \Big(\frac{6}{d^2}  + \frac{27\textup{SSR}^2}{2k^2d^2}\Big)
    \end{equation*}
    }
    under the assumption  $\me [ x^\top A x \vert \mathcal{A}\cap \mathcal{B}^c ] \ge 0$.
It is clear that when $\me [ x^\top A x \vert \mathcal{A}\cap \mathcal{B}^c ] < 0$, the above relationship also holds. 

    Finally, combining all bounds above, we obtain that $\mp(\mathcal{A}) \cdot \me \left[ x^\top A x \vert \mathcal{A} \right]$ is upper bounded by
        {\setlength{\abovedisplayskip}{2pt}
  \setlength{\belowdisplayskip}{0pt}
			\begin{align*}
				& \quad C \log{d} \cdot \mp(\mathcal{A}\cap \mathcal{B} )  
				\Big(1 + \frac{9\textup{SSR}^2}{4k} \Big)  \me [ z^\top A z \vert \mathcal{A}\cap \mathcal{B} ] \\
				& +  (x^*)^\top A x^* \Big(\frac{6}{d^2}  + \frac{27\textup{SSR}^2}{2k^2d^2}\Big).
			\end{align*}
			}
        Hence, we see that $\tr(AW^*)$ is upper bounded by the sum of $\exp\{-ck\}\me \left[ x^\top x \vert \mathcal{A}^c \right]$ and the above quantity. 
        We are now ready to show the two parts in the statement of \cref{thm:spca rand alg formal}.
        
		{\bf (Proof for part 1)}
		We write $\delta:=\exp\{-ck + 2\log{2d/k}\}$.
		Direct calculation shows that
        {\setlength{\abovedisplayskip}{2pt}
  \setlength{\belowdisplayskip}{0pt}
		\begin{align*}
			\me \left[ x^\top x \vert \mathcal{A}^c \right]
			& = \me \left[ \sum_{i: p_i > 0} \frac{(u_i^\top g)^2}{p_i^2} \epsilon_i \Bigg\vert \mathcal{A}^c \right]
			\le \me \left[ \sum_{i: p_i > 0} \frac{(u_i^\top g)^2}{p_i^2}  \Bigg\vert \mathcal{A}^c \right] \notag
            =  \sum_{i: p_i > 0} \frac{\me (u_i^\top g)^2}{p_i^2}\\
			&\le  \sum_{i: p_i = 1} \frac{\me (u_i^\top g)^2}{p_i^2} + \sum_{i: 0 < p_i < 1} \frac{\me (u_i^\top g)^2}{p_i^2}\notag
            \le  1 + \frac{9d}{4k^2} \textup{SSR}^2.
		\end{align*}
        }
		By the definition of $\delta$, one obtains that $\exp\{-ck\} \me \left[ x^\top x \vert \mathcal{A}^c \right] \le \delta$.
		
		{\bf (Proof for part 2)}
		For this part, we assume that $A\succeq 0$.
		In this special case, we claim that $1 = \norm{A} \le \frac{d}{k} \cdot (x^*)^\top A x^*$.
		This inequality holds true for the following reasoning:
		\begin{itemize}
			\item first, there exists a $(d-1)\times (d-1)$ principal submatrix $\tilde A$ of $A$ such that $ \norm{A} \le d / (d - 1) \cdot \norm{\tilde A}$ (for a reference, see, e.g., the arguments on page 189 in \cite{horn1985matrix});
			\item then, using the above fact repeatedly for the existence of a principal submatrix of size $d-2, d-3 \ldots, k$, one can obtain a principal submatrix $\hat A$ of size $k$ such that $\norm{A} \le d / k \cdot \norm{\hat A}$.
		\end{itemize}
		Combining the above inequality and the inequality in the proof for part 1, one obtains that 
        {\setlength{\abovedisplayskip}{2pt}
  \setlength{\belowdisplayskip}{0pt}
		\begin{align*}
			& \quad \exp\{-ck\} \me \left[ x^\top x \vert \mathcal{A}^c \right]
			 = \exp\{-ck\} \me \left[ x^\top x \vert \mathcal{A}^c \right] \cdot \norm{A}\\
			&\le \exp\{-ck\} \cdot \left[ 1 + \frac{9d}{4k^2} \textup{SSR}^2 \right] \cdot \frac{d}{k} \cdot (x^*)^\top A x^*
            \le \mathcal{O}\left(\frac{1}{\log{d}}\right) \cdot (x^*)^\top A x^*,
		\end{align*} 
        }
		where the last inequality comes from the assumption that $ck\ge 3\log{d/k} + \log\log{d}$. 
\end{prfc}
	
    \subsection{Proofs in \cref{sec:rank_one}}
    \label{sec:proof_rank_one}
    In this section, we prove \cref{thm:rank_one_input,prop:rank_one_exact_sign,thm:rank-one} in \cref{sec:rank_one}.
    To prove them, we need several technical lemmas. 
    We first introduce the core lemma, \cref{lem:KKT}, which is derived directly from the well-known KKT conditions~\cite{kuhn2014nonlinear}.

\begin{lemma}
	\label{lem:KKT}
	$W^* = w^* (w^*)^\top$ is an optimal solution to \ref{prob SDP SPCA} if the following conditions hold:
	\begin{enumerate}[label=B\arabic*, ref=B\arabic*]
		\item (Primal feasibility) $\onorm{w^*}\le \sqrt{k}$ and $\norm{w^*} = 1$; \label{item:primal_feasibility}
		\item (Dual feasibility) $\mu^* \ge 0$, $\lambda^* \in \R$, $Z^* \in [-1, 1]^{d\times d}$ such that $Z_{ij}^* = \sign(w_i^*) \cdot \sign(w_j^*)$ if $w_i^* w_j^* \ne 0$, and $\lambda^* I_d \succeq A - \mu^* Z^*$; \label{item:dual_feasibility}
		\item (Complementary slackness) $\mu^* (\onorm{w^*} - \sqrt{k}) = 0$, $(\lambda^* I_d - A + \mu^* Z^*) w^* = 0$. \label{item:complementary_slackness}
	\end{enumerate}
	Moreover, if $\rk(\lambda^* I_d - A + \mu^* U^*) = d - 1$, then $W^*$ is the unique optimal solution to \ref{prob SDP SPCA}.
\end{lemma}

\begin{prf}
	It is clear that the dual problem to \ref{prob SDP SPCA} is given by 
    {\setlength{\abovedisplayskip}{0pt}
  \setlength{\belowdisplayskip}{0pt}
	\begin{align*}
		\min_{\mu, \lambda, S, U} k\mu + \lambda  \qquad \text{s.t. } \mu \ge 0, \ S\succeq 0, \ U = A + S - \lambda I_d, \ \maxnorm{U}\le \mu,
	\end{align*}
    }
	and hence conditions~\ref{item:primal_feasibility}, \ref{item:dual_feasibility}, and \ref{item:complementary_slackness} are clear from KKT conditions~\cite{kuhn2014nonlinear}.
	Finally, if $\rk(\lambda^* I_d - A + \mu^* U^*) = d - 1$, we know that for any optimal solution $W_0$ to \ref{prob SDP SPCA}, it must hold that $\rk(W_0) = d - \rk(\lambda^* I_d - A + \mu^* U^*) = 1$ since $\tr(W_0 (\lambda^* I_d - A + \mu^* U^*)) = 0$.
	Due to the fact that $(\lambda^* I_d - A + \mu^* U^*) w^* = 0$, we see that $W_0$ must be equal to $w^* (w^*)^\top$.
    \end{prf}
We are now ready to prove \cref{thm:rank_one_input}.
\begin{prfc}[of \cref{thm:rank_one_input}]
    	WLOG, we assume that $\norm{u} = 1$, as a scaling of $A$ by a non-negative constant would not change its optimal solution to \ref{prob SDP SPCA}.
    	Furthermore, we assume WLOG that $\lambda = 0$, as adding a scaling of an identity matrix to $A$ would not change its optimal solution to \ref{prob SDP SPCA} either.
    	We complete the proof by utilizing \cref{lem:KKT}.
    	
    	\emph{(Primal feasibility).} For simplicity, we define $t_0:= \min_{i\in T} \abs{u_i}$, with $T=\supp(u)$. 
    	Since $\onorm{u} > \sqrt{k}$, it is clear that $|T| \ge \onorm{u}^2 > k$.
    	We define a unit vector $w(t):= (u - t \sign(u)) / \norm{u - t \sign(u)}\in \R^d$ with $t\in [0, t_0]$. 
        We claim that $\onorm{w(t_0)} < \sqrt{k}$. 
    	Indeed, the claim is equivalent to the quadratic inequality $|T|(|T| - k) t_0^2 - 2\onorm{u}(|T| - k) t_0 + (\onorm{u}^2 - k) < 0$, which is then implied by \eqref{eqn:ass rank one}.
		Therefore, by the Intermediate Value Theorem, there exists $t^*\in (0, t_0)$ such that $\onorm{w(t^*)} = \sqrt{k}$.
		
		We thus define $w^* = w(t^*)$, it is clear that $\norm{w^*} = 1$, $\onorm{w^*} = \sqrt{k}$, and $\supp(w^*) = T$.
    	
    	\emph{(Dual feasibility).} By the definition of $w^*$, we see that $\sign(w^*) = \sign(u)$, and we obtain that
        {\setlength{\abovedisplayskip}{0pt}
  \setlength{\belowdisplayskip}{0pt}
    	\begin{align*}
    		u = \norm{u - t^*\sign(u)} w^* + t^* \sign(w^*) := \alpha w^* + \beta \sign(w^*),
    	\end{align*}
        }
    	where $\alpha, \beta > 0$ by definition. 
        We then define
        {\setlength{\abovedisplayskip}{0pt}
  \setlength{\belowdisplayskip}{0pt}
    	\begin{align*}
    		\lambda^* &:= \norm{u - t^*\sign(u)} (u^\top w^*) = \alpha (u^\top w^*) = \alpha (\alpha + \beta \sqrt{k}) > 0\\
    		\mu^* &:= \frac{t^* (u^\top w^*)}{\sqrt{k}} = \frac{\beta (u^\top w^*)}{\sqrt{k}} = \frac{\beta (\alpha + \beta \sqrt{k})}{\sqrt{k}} > 0.
    	\end{align*}
        }
    	We further define $Z^* = \sign(u) \sign(u)^\top$, and we show that 
        {\setlength{\abovedisplayskip}{0pt}
  \setlength{\belowdisplayskip}{0pt}
    	\begin{align*}
    		\lambda^* I_d \succeq uu^\top - \mu^* Z^* 
    		\Longleftrightarrow 
    			\lambda^* I_d \succeq uu^\top - \mu^* \sign(u) \sign(u)^\top.
    	\end{align*}
        }
    	For simplicity, define $M:= \lambda^* I_d - uu^\top + \mu^* \sign(u) \sign(u)^\top$.
    	Since $u\in \sp1(w^*, \sign(w^*))$, it is clear that for any non-zero vector $v\in \sp1(w^*, \sign(w^*))^\perp$, $v^\top M v = \lambda^* \norm{v}^2 > 0$.
    	Next, take any non-zero vector $v\in \sp1(w^*, \sign(w^*))$, we also show that $v^\top M v \ge 0$.
    	First, we observe that 
        {\setlength{\abovedisplayskip}{0pt}
  \setlength{\belowdisplayskip}{0pt}
    	\begin{align*}
    		M w^* = (\lambda^* I_d - uu^\top + \mu^* \sign(u) \sign(u)^\top) w^*
    		= \lambda^* w^* - (u^\top w^*) u + \mu^* \sqrt{k} \sign(u)
    		 = 0,
    	\end{align*}
        }
    	where the last equality follows from the fact that $u = \alpha w^* + \beta \sign(w^*) = \alpha w^* + \beta \sign(u)$ and how $\lambda^*$ and $\mu^*$ are defined.
        For simplicity, define $s:=\sign(w^*) = \sign(u)$, and we notice that
        {\setlength{\abovedisplayskip}{0pt}
  \setlength{\belowdisplayskip}{0pt}
    	\begin{align*}
    	s^\top M s	
    	&= s^\top \left(\lambda^* I_d - uu^\top + \mu^* \sign(u) \sign(u)^\top\right) s
        = (\lambda^* + \mu^* |T|) |T| - \onorm{u}^2\\
	    		 &= \alpha^2|T| + \alpha\beta\sqrt{k}|T| + \frac{\alpha\beta|T|^2}{\sqrt{k}} + \beta^2 |T|^2 - (\alpha\sqrt{k} + \beta |T|)^2 \\
	    		 &= \alpha^2 (|T| - k) + \frac{\alpha\beta|T|}{\sqrt{k}} (|T| - k) > 0.
	    	\end{align*}
	        }
    	Thus, for any $v = a w^* + b\sign(w^*)$, we see that
    	$v^\top M v = b^2 \sign(w^*)^\top M \sign(w^*) \ge 0$.
    	
    	 Hence, we have shown that $M\succeq 0$, with its smallest eigenvalue being zero, and a positive second smallest eigenvalue. 
  		This implies that $\rk(M) = d - 1$.
    	
    	\emph{(Complementary slackness).} The complementary slackness conditions follow from the fact that $\onorm{w^*} = \sqrt{k}$ and $Mw^* = 0$. 
    \end{prfc}

    We now turn to \cref{prop:rank_one_exact_sign,thm:rank-one}. We first state the low-level certificate, \cref{prop:rank_one_resolvent}, that underlies both results. We then prove \cref{prop:rank_one_exact_sign}, establish the supporting lemmas behind Assumption~\ref{ass:rank-one}, and finally deduce \cref{thm:rank-one}.

\begin{proposition}
\label{prop:rank_one_resolvent}
Assume $A\succeq 0$ and let $S\subseteq[d]$ be nonempty.
Let $A_{S,S}$ have eigenvalues $\lambda_1>\lambda_2\ge 0$.
Fix a sign vector $s\in\{\pm 1\}^{|S|}$ and suppose there exists
$\lambda^*\in(\lambda_2,\lambda_1)$ such that, with
$w := (A_{S,S}-\lambda^* I)^{-1} s$,
\begin{enumerate}[label=Q\arabic*, ref=Q\arabic*]
  \item \label{item:resolvent_sign}
  the vector $w$ satisfies $\sign(w)=s$;
  \item \label{item:resolvent_level}
  $\|w\|_1/\|w\|_2 = \sqrt{k}$;
  \item \label{item:resolvent_offsupport}
  with $\mu^* := 1/(\sqrt{k}\,\|w\|_2)$, one has
  $\max\{\max_{i\in S, j\in S^c}|A_{ij}|,\ \max_{i,j\in S^c}|A_{ij}|\}\le \mu^*$;
  \item \label{item:resolvent_gap}
  $\lambda^* - \lambda_2 \ge \mu^* |S|$.
\end{enumerate}
Then \ref{prob SDP SPCA} admits a rank-one optimal solution
$W^* = w^*(w^*)^\top$, where $w^*\in\R^d$ is defined by
{\setlength{\abovedisplayskip}{0pt}
  \setlength{\belowdisplayskip}{0pt}
\begin{equation*}
  (w^*)_S = \frac{w}{\|w\|_2},
  \qquad
  (w^*)_{S^c}=0.
\end{equation*}
}
Moreover, if \ref{item:resolvent_gap} is strict, then $W^*$ is the unique
optimal solution.
\end{proposition}

\begin{prf}
We verify the KKT conditions in \cref{lem:KKT}. By construction, $w^*$ is
supported on $S$ and satisfies $\|w^*\|_2=1$ and, by
\ref{item:resolvent_level}, $\|w^*\|_1=\sqrt{k}$.
Thus $w^*$ is primal feasible.

Set $\mu^*:=1/(\sqrt{k}\,\|w\|_2)$. Let $Z^*\in[-1,1]^{d\times d}$ be
{\setlength{\abovedisplayskip}{0pt}
  \setlength{\belowdisplayskip}{0pt}
\begin{equation*}
  Z^*_{S,S}:= s s^\top,
  \qquad
  Z^*_{ij}:=A_{ij}/\mu^*\ \text{for } (i,j)\notin S\times S.
\end{equation*}
}
By \ref{item:resolvent_offsupport}, $|Z^*_{ij}|\le 1$ for all $(i,j)$, and by
\ref{item:resolvent_sign} we have $\sign(w^*_S)=s$, so $Z^*_{S,S}$ matches the
sign pattern on the support block.

	We now show $\lambda^* I_d \succeq A-\mu^* Z^*$. Since
	$A-\mu^* Z^*$ has zero entries outside $S\times S$, it suffices to show
  {\setlength{\abovedisplayskip}{0pt}
  \setlength{\belowdisplayskip}{0pt}
\begin{equation*}
  \lambda^* I_{|S|} \succeq A_{S,S} - \mu^* s s^\top.
	\end{equation*}
  }
	Let $M:=\lambda^* I_{|S|} - A_{S,S} + \mu^* s s^\top$.
	Using $(A_{S,S}-\lambda^* I_{|S|}) w = s$, we obtain
  {\setlength{\abovedisplayskip}{0pt}
  \setlength{\belowdisplayskip}{0pt}
	\begin{equation*}
	  M w = 0.
	\end{equation*}
  }
	Let $H:=\lambda^* I_{|S|} - A_{S,S}$.
  The eigenvalues of $H$ are $\lambda^* - \lambda_i(A_{S,S})$, so $H$ has exactly one negative eigenvalue and $|S|-1$ positive eigenvalues because $\lambda_1>\lambda^*>\lambda_2$.
	Since $M = H + \mu^* s s^\top$ is a rank-one positive-semidefinite update of $H$, Weyl's inequality implies that $M$ has at most one nonpositive eigenvalue.
	Because $Mw=0$ and $w\neq 0$, zero is an eigenvalue of $M$, so $M$ cannot have a negative eigenvalue. 
  Hence $M\succeq 0$, and dual feasibility holds.

Complementary slackness is satisfied because
$\mu^*(\|w^*\|_1-\sqrt{k})=0$ and
{\setlength{\abovedisplayskip}{0pt}
  \setlength{\belowdisplayskip}{0pt}
\begin{equation*}
  (\lambda^* I_d - A + \mu^* Z^*) w^* = 0,
\end{equation*}
}
	which follows from $M w = 0$ and the zero off-support blocks. 
  Therefore, $W^*=w^*(w^*)^\top$ is optimal by \cref{lem:KKT}. 
  If \ref{item:resolvent_gap} is strict, then Weyl's inequality yields that the second smallest eigenvalue of $M$ is strictly positive, so $\rk(\lambda^* I_d - A + \mu^* Z^*)=d-1$ and uniqueness follows from \cref{lem:KKT}.
\end{prf}

We show now how this proposition can be used to prove \cref{prop:rank_one_exact_sign}:

\begin{prfc}[of \cref{prop:rank_one_exact_sign}]
Let
{\setlength{\abovedisplayskip}{0pt}
  \setlength{\belowdisplayskip}{0pt}
\begin{equation*}
  \eta:=\max\{\max_{i\in S, j\in S^c}|A_{ij}|, \max_{i,j\in S^c}|A_{ij}|\}.
\end{equation*}
}
If $\eta<(\lambda_1-\lambda_2)/(2k)$, choose any $\lambda^*\in\left((\lambda_1+\lambda_2)/2,\lambda_1-k\eta\right)$, otherwise $\eta=(\lambda_1-\lambda_2)/(2k)$, and we set $\lambda^*:=(\lambda_1+\lambda_2)/2$.
In either case, $\lambda^*\in(\lambda_2,\lambda_1)$ and
{\setlength{\abovedisplayskip}{0pt}
  \setlength{\belowdisplayskip}{0pt}
\begin{equation*}
  \eta\le \frac{\lambda_1-\lambda^*}{k},
  \qquad
  \lambda^* - \lambda_2\ge \lambda_1-\lambda^*,
\end{equation*}
}
with strict inequality in the second relation when \eqref{eqn:off-support_upper_bound} is strict.

Now set $w:=(B-\lambda^* I)^{-1}s$.
Since $s=\sqrt{k}\,v_1$ lies entirely in the $\lambda_1$-eigenspace,
{\setlength{\abovedisplayskip}{0pt}
  \setlength{\belowdisplayskip}{0pt}
\begin{equation*}
  w=\frac{s}{\lambda_1-\lambda^*}.
\end{equation*}
}
Therefore $\sign(w)=s$ and
{\setlength{\abovedisplayskip}{0pt}
  \setlength{\belowdisplayskip}{0pt}
\begin{equation*}
  \frac{\|w\|_1}{\|w\|_2}
  =\frac{\|s\|_1}{\|s\|_2}
  =\sqrt{k}.
\end{equation*}
}
Moreover,
{\setlength{\abovedisplayskip}{0pt}
  \setlength{\belowdisplayskip}{0pt}
\begin{equation*}
  \mu^*:=\frac{1}{\sqrt{k}\,\|w\|_2}
  =\frac{\lambda_1-\lambda^*}{k},
\end{equation*}
}
so the choice of $\lambda^*$ gives $\eta\le \mu^*$ and
{\setlength{\abovedisplayskip}{0pt}
  \setlength{\belowdisplayskip}{0pt}
\begin{equation*}
  \lambda^* - \lambda_2\ge \mu^* k=\mu^*|S|,
\end{equation*}
}
again with strict inequality in the last inequality when the hypothesis is strict. 
Thus \ref{item:resolvent_sign}--\ref{item:resolvent_gap} in \cref{prop:rank_one_resolvent} all hold. 
Applying \cref{prop:rank_one_resolvent} yields the claimed optimal solution. 
Since $w/\|w\|_2=s/\sqrt{k}$, the resulting vector $w^*$ satisfies
{\setlength{\abovedisplayskip}{0pt}
  \setlength{\belowdisplayskip}{0pt}
\begin{equation*}
  (w^*)_S=\frac{s}{\sqrt{k}},
  \qquad
  (w^*)_{S^c}=0.
\end{equation*}
}
The strict case gives uniqueness by the last part of \cref{prop:rank_one_resolvent}.
\end{prfc}

In the remainder of this subsection, we prove \cref{thm:rank-one}.
We first show how the items in Assumption~\ref{ass:rank-one} translate to the hypotheses of \cref{prop:rank_one_resolvent}.

\begin{lemma}
\label{lem:sign_preservation}
Let $A\in\R^{d\times d}$ be symmetric with eigenvalues
$\lambda_1>\lambda_2\ge\cdots\ge\lambda_d$.
Let $v_1$ be a unit eigenvector associated with $\lambda_1$, assume that $v_1$
has no zero coordinates, and define
$s := \sign(v_1)\in\{\pm1\}^d$ and
$m := \min_{1\le i\le d}|(v_1)_i|>0$.
Let $P_2$ denote the orthogonal projection onto the eigenspace associated with
$\lambda_2$, and for each eigenvalue $\theta<\lambda_2$, let $P_\theta$ denote
the orthogonal projection onto the $\theta$-eigenspace. Assume:
\begin{enumerate}[label=(H\arabic*), leftmargin=2.5em]
  \item $P_2 s=0$;
  \item $\sum_{\theta<\lambda_2} \frac{\norm{P_\theta s}^2}{(\lambda_2-\theta)^2}
    < \left(\frac{\onorm{v_1}\,m}{\lambda_1-\lambda_2}\right)^2$.
\end{enumerate}
Then
{\setlength{\abovedisplayskip}{0pt}
  \setlength{\belowdisplayskip}{0pt}
\begin{equation*}
  \sign\big((A-\lambda I)^{-1}s\big)=s,
  \qquad \forall \lambda\in(\lambda_2,\lambda_1).
\end{equation*}
}
\end{lemma}

\begin{prf}
Fix $\lambda\in(\lambda_2,\lambda_1)$.
Since $A$ is symmetric, its spectral decomposition can be written as
{\setlength{\abovedisplayskip}{0pt}
  \setlength{\belowdisplayskip}{2pt}
\begin{equation*}
  A=\lambda_1 v_1 v_1^\top + \lambda_2 P_2 + \sum_{\theta<\lambda_2}\theta P_\theta,
\end{equation*}
}
where the summation runs over the distinct eigenvalues below $\lambda_2$. Hence
\begin{equation*}
  (A-\lambda I)^{-1}s
  = \frac{v_1^\top s}{\lambda_1-\lambda}v_1
    + \frac{P_2 s}{\lambda_2-\lambda}
    + \sum_{\theta<\lambda_2}\frac{P_\theta s}{\theta-\lambda}.
\end{equation*}
Since $s=\sign(v_1)$, one has $v_1^\top s=\onorm{v_1}$; by
assumption~\textup{(H1)}, the $\lambda_2$-pole vanishes. Therefore
\begin{equation*}
  (A-\lambda I)^{-1}s
  = \frac{\onorm{v_1}}{\lambda_1-\lambda}v_1 + r(\lambda), \textup{ where } r(\lambda):=\sum_{\theta<\lambda_2}\frac{P_\theta s}{\theta-\lambda}.
\end{equation*}
By orthogonality of the spectral subspaces,
\begin{equation*}
  \norm{r(\lambda)}^2
  = \sum_{\theta<\lambda_2}\frac{\norm{P_\theta s}^2}{(\theta-\lambda)^2}
  \le \sum_{\theta<\lambda_2}\frac{\norm{P_\theta s}^2}{(\lambda_2-\theta)^2}
  =: M^2,
\end{equation*}
because $|\theta-\lambda|=\lambda-\theta>\lambda_2-\theta$ whenever
$\theta<\lambda_2<\lambda$. Hence
\begin{equation*}
  \|r(\lambda)\|_\infty\le \norm{r(\lambda)}\le M.
\end{equation*}
Assumption~\textup{(H2)} gives
\begin{equation*}
  M<\frac{\onorm{v_1}\,m}{\lambda_1-\lambda_2}
  <\frac{\onorm{v_1}\,m}{\lambda_1-\lambda},
\end{equation*}
since $\lambda\in(\lambda_2,\lambda_1)$. Therefore, for each coordinate $i$,
\begin{equation*}
  s_i\big((A-\lambda I)^{-1}s\big)_i
  = \frac{\onorm{v_1}}{\lambda_1-\lambda}s_i(v_1)_i + s_i r_i(\lambda)
  \ge \frac{\onorm{v_1}\,m}{\lambda_1-\lambda} - \|r(\lambda)\|_\infty
  > 0,
\end{equation*}
because $s_i=\sign((v_1)_i)$. Thus each coordinate of $(A-\lambda I)^{-1}s$
has sign $s_i$, as desired.
\end{prf}

The next lemma records the only endpoint limit that we need later,
namely the behavior as $\lambda\uparrow\lambda_1$.

\begin{lemma}
\label{lem:shifted_operator_limit}
Let $A\in\R^{d\times d}$ be symmetric and positive semi-definite with spectral decomposition
\begin{equation*}
  A=\lambda_1 v_1 v_1^\top + \lambda_2 P_2 + \sum_{\theta<\lambda_2}\theta P_\theta,
\end{equation*}
where $\lambda_1>\lambda_2\ge\cdots\ge 0$, $v_1$ is a unit eigenvector for $\lambda_1$. 
Fix $s\in\{\pm1\}^d$, and define
\begin{equation*}
  w(\lambda):=(A-\lambda I)^{-1}s,
  \qquad
  f(\lambda):=\frac{\|w(\lambda)\|_1}{\|w(\lambda)\|_2},
  \qquad \lambda\in(\lambda_2,\lambda_1).
\end{equation*}
If $v_1^\top s\neq 0$, then
{\setlength{\abovedisplayskip}{0pt}
  \setlength{\belowdisplayskip}{0pt}
\begin{equation*}
  \lim_{\lambda\uparrow\lambda_1} f(\lambda)=\frac{\|v_1\|_1}{\|v_1\|_2}.
\end{equation*}
}
\end{lemma}

\begin{prf}
We borrow the notation in Assumption~\ref{ass:rank-one}.
Let $P_2$ be the orthogonal projection onto the $\lambda_2$-eigenspace, and each $P_\theta$ is the orthogonal projection onto the $\theta$-eigenspace for $\theta<\lambda_2$.
By the spectral decomposition,
\begin{equation*}
  w(\lambda)
  = \frac{v_1^\top s}{\lambda_1-\lambda}v_1
    + \frac{P_2 s}{\lambda_2-\lambda}
    + \sum_{\theta<\lambda_2}\frac{P_\theta s}{\theta-\lambda},
\end{equation*}
for every $\lambda\in(\lambda_2,\lambda_1)$. 
Note that $f(\lambda)$ is scale-invariant: multiplying $w(\lambda)$ by any nonzero scalar does not change its value.
Then, multiplying the expansion above by $\lambda_1-\lambda$ gives
\begin{equation*}
  (\lambda_1-\lambda)w(\lambda)
  = (v_1^\top s)v_1
    + \frac{\lambda_1-\lambda}{\lambda_2-\lambda}P_2 s
    + \sum_{\theta<\lambda_2}\frac{\lambda_1-\lambda}{\theta-\lambda}P_\theta s.
\end{equation*}
As $\lambda\uparrow\lambda_1$, the second and third terms vanish, so
$(\lambda_1-\lambda)w(\lambda)\to (v_1^\top s)v_1$. By scale invariance,
\begin{equation*}
  \lim_{\lambda\uparrow\lambda_1} f(\lambda)
  = \frac{\|(v_1^\top s)v_1\|_1}{\|(v_1^\top s)v_1\|_2}
  = \frac{\|v_1\|_1}{\|v_1\|_2}.
\end{equation*}
\end{prf}

It remains to control the scalar spectral-margin condition
\ref{item:resolvent_gap} in the general case.

\begin{lemma}
\label{lem:q4_quadratic}
With the notation of \cref{prop:rank_one_resolvent}, let $B:=A_{S,S}$ and assume that $B$ and $s$ satisfy the hypotheses of \cref{lem:sign_preservation}.
For $\lambda\in(\lambda_2,\lambda_1)$, define
\begin{equation*}
  w(\lambda):=(B-\lambda I)^{-1}s,
  \qquad
  \alpha:=\onorm{v_1}\,m,
  \qquad
  M:=\left(\sum_{\theta<\lambda_2}\frac{\norm{P_\theta s}^2}{(\lambda_2-\theta)^2}\right)^{1/2},
  \qquad
  \Delta:=\lambda_1-\lambda_2.
\end{equation*}
If $\lambda\in(\lambda_2,\lambda_1)$ satisfies \ref{item:resolvent_level} and
\begin{equation*}
  (\lambda-\lambda_2)\left(\frac{\alpha}{\lambda_1-\lambda}-M\right)\ge 1,
\end{equation*}
then \ref{item:resolvent_gap} holds at that $\lambda$. 
Equivalently, writing $x:=\lambda_1-\lambda$, this sufficient condition is
\begin{equation*}
  Mx^2-(\Delta M+\alpha+1)x+\Delta\alpha\ge 0.
\end{equation*}
In particular, there exists $\varepsilon>0$ such that every $\lambda\in(\lambda_1-\varepsilon,\lambda_1)$ satisfying \ref{item:resolvent_level} also satisfies \ref{item:resolvent_gap}.
\end{lemma}

\begin{prf}
Fix $\lambda\in(\lambda_2,\lambda_1)$ and set $w:=w(\lambda)$. The proof of
\cref{lem:sign_preservation} yields, for each coordinate,
\begin{equation*}
  s_i w_i\ge \frac{\alpha}{\lambda_1-\lambda}-M.
\end{equation*}
Since the same lemma gives $\sign(w)=s$, we have $s_i w_i=|w_i|$, hence
\begin{equation*}
  \|w\|_1=\sum_{i=1}^{|S|}|w_i|=\sum_{i=1}^{|S|} s_i w_i
  \ge |S|\left(\frac{\alpha}{\lambda_1-\lambda}-M\right).
\end{equation*}
If \ref{item:resolvent_level} holds at $\lambda$, then
\begin{equation*}
  \mu(\lambda)=\frac{1}{\sqrt{k}\,\|w\|_2}=\frac{1}{\|w\|_1}.
\end{equation*}
Therefore,
\begin{equation*}
  (\lambda-\lambda_2)\|w\|_1
  \ge |S|(\lambda-\lambda_2)\left(\frac{\alpha}{\lambda_1-\lambda}-M\right)
  \ge |S|,
\end{equation*}
which is exactly \ref{item:resolvent_gap}.

Now write $x:=\lambda_1-\lambda$, so that $\lambda-\lambda_2=\Delta-x$. Then
\begin{equation*}
  (\lambda-\lambda_2)\left(\frac{\alpha}{\lambda_1-\lambda}-M\right)\ge 1
  \iff (\Delta-x)\left(\frac{\alpha}{x}-M\right)\ge 1.
\end{equation*}
Since $x>0$, multiplying by $x$ gives
\begin{equation*}
  \alpha\Delta-\alpha x-M\Delta x+Mx^2\ge x,
\end{equation*}
which is equivalent to
\begin{equation*}
  Mx^2-(\Delta M+\alpha+1)x+\Delta\alpha\ge 0.
\end{equation*}
Finally, the left-hand side at $x=0$ equals $\Delta\alpha>0$, so by continuity it remains nonnegative for all sufficiently small $x>0$. Equivalently, there exists $\varepsilon>0$ such that the inequality holds whenever $\lambda\in(\lambda_1-\varepsilon,\lambda_1)$, and the last claim follows.
\end{prf}

Finally, we deduce the high-level theorem from the packaged assumptions.

\begin{prfc}[of \cref{thm:rank-one}]
Set
\begin{equation*}
  R(\lambda):=\frac{\|(A_{S,S}-\lambda I_{|S|})^{-1}s\|_1}{\|(A_{S,S}-\lambda I_{|S|})^{-1}s\|_2},
  \qquad \lambda\in(\lambda_2,\lambda_1).
\end{equation*}
By \ref{item:ass_resolvent_sign} and \cref{lem:sign_preservation}, applied to $B=A_{S,S}$, one has
\begin{equation*}
  \sign\big((A_{S,S}-\lambda I_{|S|})^{-1}s\big)=s,
  \qquad \forall \lambda\in(\lambda_2,\lambda_1),
\end{equation*}
so \ref{item:resolvent_sign} holds throughout $(\lambda_2,\lambda_1)$. 
Moreover, the proof of \cref{lem:sign_preservation} yields
\begin{equation*}
  (A_{S,S}-\lambda I_{|S|})^{-1}s
  = \frac{\onorm{v_1}}{\lambda_1-\lambda}v_1
    + \sum_{\theta<\lambda_2}\frac{P_\theta s}{\theta-\lambda},
\end{equation*}
because \ref{item:ass_resolvent_sign} gives $P_2 s=0$. 
Letting $\lambda\downarrow\lambda_2$ therefore shows
\begin{equation*}
  \lim_{\lambda\downarrow\lambda_2} R(\lambda)=\frac{\|w_2\|_1}{\|w_2\|_2}.
\end{equation*}
Since $v_1^\top s=\onorm{v_1}\neq 0$, \cref{lem:shifted_operator_limit},
applied to $B=A_{S,S}$, also gives
\begin{equation*}
  \lim_{\lambda\uparrow\lambda_1} R(\lambda)=\frac{\|v_1\|_1}{\|v_1\|_2}.
\end{equation*}
The map $\lambda\mapsto (A_{S,S}-\lambda I_{|S|})^{-1}s$ is continuous and nonzero on $(\lambda_2,\lambda_1)$, so $R$ is continuous there. 
Hence \ref{item:ass_resolvent_endpoints} and the intermediate value theorem imply that the equation $R(\lambda)=\sqrt{k}$ has a solution in $(\lambda_2,\lambda_1)$. 
Let $\lambda^*\in(\lambda_2,\lambda_1)$ be the value from \ref{item:ass_resolvent_window}, and set
\begin{equation*}
  w := (A_{S,S}-\lambda^* I_{|S|})^{-1}s.
\end{equation*}
Then \ref{item:resolvent_sign} holds by our previous discussion, and the first assertion in \ref{item:ass_resolvent_window} gives $\|w\|_1/\|w\|_2=\sqrt{k}$, hence \ref{item:resolvent_level}. 
The second assertion in \ref{item:ass_resolvent_window} and \cref{lem:q4_quadratic} imply \ref{item:resolvent_gap} at $\lambda^*$. 
Finally, the off-support bound in \ref{item:ass_resolvent_offsupport} is exactly \ref{item:resolvent_offsupport}.

All four hypotheses of \cref{prop:rank_one_resolvent} are therefore satisfied, so \ref{prob SDP SPCA} admits a rank-one optimal solution $W^*=w^*(w^*)^\top$. 
Since \ref{item:resolvent_sign} gives $\sign(w)=s\in\{\pm1\}^{|S|}$, every coordinate of $w$ on $S$ is nonzero, and hence $\supp(w^*)=S$.

If the spectral-margin inequality in \ref{item:ass_resolvent_window} is strict, then the proof of \cref{lem:q4_quadratic} yields strict \ref{item:resolvent_gap}. 
Uniqueness therefore follows from \cref{prop:rank_one_resolvent}.
\end{prfc}

\section{Proofs in \cref{sec:adversarial pert}}
	\label{sec:proof of adver}
	In this section, we present the proofs of \cref{prop:robust random,thm:randomized_alg_in_stat_model} in \cref{sec:adversarial pert}.

    We first provide the proof of \cref{prop:robust random}.
    In order to do so, we need several propositions.     
    The first proposition characterizes how close $W^*$ is to the ``sparse spike'' in a deterministic model: 
	
	\begin{proposition}
		\label{prop:robust deter}
		Suppose the input matrix in \ref{prob SPCA} can be written as $A = B^\top B + E$, where $B^\top B$ admits a $k$-sparse eigenvector $v$ corresponding to its largest eigenvalue $\lambda_1(B^\top B)$, and $E\in \R^{d\times d}$ is a matrix such that $\infnorm{E} \le a$. 
		Assume that $\lambda_1(B^\top B) - \lambda_2(B^\top B) > 0$, and denote $W^*$ an optimal solution to \ref{prob SDP SPCA}. 
		Then, we have 
		{\setlength{\abovedisplayskip}{2pt}
  \setlength{\belowdisplayskip}{0pt}
        \begin{align*}
			\|W^* - v v^\top\|_F \le \frac{2ak}{\lambda_1 - \lambda_2}  + \sqrt{\frac{2ak}{\lambda_1 - \lambda_2}}
		\end{align*}
        }
	\end{proposition}

 To prove \cref{prop:robust deter}, we utilize the curvature lemma presented in \cite{vu2013fantope}. 
	This lemma helps transform the problem of bounding the Frobenius distance of $W^*$ and $v v^\top$ into bounding the difference of their objective values for \ref{prob SDP SPCA}, which is easier to handle.
	
	\begin{lemma}[Lemma~3.1 in \cite{vu2013fantope}]
		\label{lem:curvature}
		Let $B$ be a symmetric $d\times d$ matrix and $P$ be the projection onto the subspace
		spanned by the eigenvectors of $B$ corresponding to its $l$ largest eigenvalues $\lambda_1 \ge \lambda_2 \ge \cdots \ge \lambda_l$. If $\delta:=\lambda_l - \lambda_{l+1} > 0$, then for any symmetric matrix $F$ satisfying $0\preceq F \preceq I_d$ and $\tr(F) = l$, we have
        { \setlength{\abovedisplayskip}{2pt}
  \setlength{\belowdisplayskip}{0pt}
		\begin{align*}
			\fnorm{P - F}^2 \le \frac{2}{\delta} \tr(B (P - F)).
		\end{align*}
        }
	\end{lemma}
	\begin{remark}
		In \cref{lem:curvature}, suppose $B$ admits a singular value decomposition $B = \sum_{i = 1}^d \lambda_{i} u_i u_i^\top$, then the matrix $P=\sum_{i = 1}^l u_i u_i^\top$.
	\end{remark}
	
	\begin{prfc}[of \cref{prop:robust deter}]
		Define $G:=B^\top B$ and $S:=\supp(v)$. We denote by $\lambda_1:= \lambda_1(G)$ and $\lambda_2:=\lambda_2(G)$ the largest and second largest eigenvalues of $G$, respectively.
		By \cref{lem:curvature}, we obtain that 
		\begin{equation*}
			\begin{aligned}
				\fnorm{v v^\top - W^*}^2 
				& \le \frac{2}{\lambda_1 - \lambda_2} \tr\left(G(v v^\top - W^*)\right) 
	                = \frac{2}{\lambda_1 - \lambda_2} \tr\left( (A - E) (v v^\top - W^*) \right).
			\end{aligned}
		\end{equation*}
		Since $v v^\top$ is feasible to problem \ref{prob SDP SPCA}, we have $\tr(A W^*) \ge \tr(A v v^\top)$. Combining this with the above inequality, we see that
        {\setlength{\abovedisplayskip}{2pt}
  \setlength{\belowdisplayskip}{0pt}
		\begin{align*}
			\fnorm{v v^\top - W^*}^2
			&\le \frac{2}{\lambda_1 - \lambda_2} \left[ - \tr\Big( E (vv^\top - W^*)   \Big)  \right]
            \le \frac{2 \infnorm{E}}{\lambda_1 - \lambda_2}  \onorm{v v^\top - W^*} \le \frac{2a \onorm{v v^\top - W^*}}{\lambda_1 - \lambda_2} .
		\end{align*}
        }
		Let $\hat W^*_{S^c}$ be a matrix such that it is zero out in $(S,S)$ block, and it coincides with the remaining entries in $W^*$.
		By Cauchy-Schwartz inequality and the fact that $\onorm{\hat W^*_{S^c}} \le k$, we obtain that 
        {\setlength{\abovedisplayskip}{2pt}
  \setlength{\belowdisplayskip}{0pt}
		\begin{align*}
			\onorm{v v^\top - W^*}
			= \onorm{v_S v_S^\top - W^*_{S,S}} + \onorm{\hat W^*_{S^c}}
			\le k \fnorm{v_S v_S^\top - W^*_{S,S}} + k
		\end{align*}
        }
		Therefore, we see that
        {\setlength{\abovedisplayskip}{2pt}
  \setlength{\belowdisplayskip}{0pt}
		\begin{align*}
			\fnorm{v v^\top - W^*}^2
			\le \frac{2 a k}{\lambda_1 - \lambda_2} \fnorm{v_S v_S^\top - W^*_{S,S}} + \frac{2 a k}{\lambda_1 - \lambda_2}.
		\end{align*}
		}
        
		We conclude the proof by noticing that, for some non-negative number $c\ge 0$, the quadratic inequality $x^2 \le c x + c$ implies that $x \le (c + \sqrt{c^2 + 4c}) / 2 \le c + \sqrt{c}$.
        \end{prfc}

	To apply \cref{prop:robust deter} to \cref{model:adversarial}, we need the following high probability bound:
    \begin{lemma}
    \label{lem:inf_norm_bound_of_E}
        In \cref{model:adversarial}, denote $\lambda_1, \lambda_2$ to be the largest and second largest eigenvalue of $\Sigma$, respectively, and assume $\lambda_1 - \lambda_2 > 0$. 
        Write
        {\setlength{\abovedisplayskip}{2pt}
  \setlength{\belowdisplayskip}{0pt}
		\begin{align}
        \label{eqn:def_of_E}
			A = n \Sigma + \left(B^\top B - n \Sigma \right) +  \left(M^\top M + M^\top B + B^\top M\right) =: n\Sigma + E.
		\end{align}
        }
        Then,
        {\setlength{\abovedisplayskip}{2pt}
  \setlength{\belowdisplayskip}{0pt}
		\begin{align}
  \label{eqn:inf_norm_bound_of_E}
			\frac{1}{n}\infnorm{E} 
   \le C \sigma^2 \left(\sqrt{\frac{\log{d}}{n}} + \frac{\log{d}}{n} \right) + \frac{b^2}{n} 
   + \frac{2b}{\sqrt{n}} \cdot \sqrt{C \sigma^2 \left(\sqrt{\frac{\log{d}}{n}} + \frac{\log{d}}{n} \right)  + \max_{i\in [d]} \Sigma_{ii} }
		\end{align}
        }
        with probability at least $1 - \mathcal{O}(d^{-10})$ for some absolute constant $C > 0$.
    \end{lemma}

    \begin{prf}
        We first recall that each row of $B$ has zero-mean and admits a covariance matrix $\Sigma$ in \cref{model:adversarial}.
		As every single entry of $B^\top B$ is a summation of sub-Gaussian random variables with parameter $\sigma^2$, then by Bernstein's inequality (see, e.g., Theorem~2.8.1 in \cite{vershynin2018high}) and an argument of union bound, we obtain that
{\setlength{\abovedisplayskip}{2pt}
  \setlength{\belowdisplayskip}{0pt}
		\begin{align}
			\label{eqn:bernstein bound}
			\infnorm{\frac{1}{n}B^\top B - \Sigma} \le C \sigma^2 \left(\sqrt{\frac{\log{d}}{n}} + \frac{\log{d}}{n} \right)
		\end{align}
        }
		with probability at least $1 - d^{-10}$ for some absolute constant $C > 0$. 
		By Cauchy Schwartz inequality, we also see that
        {\setlength{\abovedisplayskip}{2pt}
  \setlength{\belowdisplayskip}{0pt}
		\begin{align}
			\label{eqn:bounds for infnorm MM}
			\infnorm{B^\top M} \le \ottnorm{B} \ottnorm{M} \le b\ottnorm{B}, \qquad \infnorm{M^\top M} \le \ottnorm{M}^2 \le b^2
		\end{align}
        }
		Then, notice that
        {\setlength{\abovedisplayskip}{2pt}
  \setlength{\belowdisplayskip}{0pt}
		\begin{align}
			\label{eqn:bounds for infnorm BM}
			\frac{1}{n}\ottnorm{B}^2 = \max_{i\in [d]} \left(\frac{1}{n} B^\top B \right)_{ii} \stackrel{\eqref{eqn:bernstein bound}}{\le} C \sigma^2 \left(\sqrt{\frac{\log{d}}{n}} + \frac{\log{d}}{n} \right) + \max_{i\in [d]} \Sigma_{ii} 
		\end{align} 
        }
		with probability at least $1 - d^{-10}$.
		Recall that we write
{\setlength{\abovedisplayskip}{2pt}
  \setlength{\belowdisplayskip}{0pt}
		\begin{align*}
			A = n \Sigma + \left(B^\top B - n \Sigma \right) +  \left(M^\top M + M^\top B + B^\top M\right) = n\Sigma + E.
		\end{align*}
        }
		Combining \eqref{eqn:bernstein bound}, \eqref{eqn:bounds for infnorm MM}, \eqref{eqn:bounds for infnorm BM}, we see that
        {\setlength{\abovedisplayskip}{2pt}
  \setlength{\belowdisplayskip}{0pt}
		\begin{align*}
			\frac{1}{n}\infnorm{E} \le C \sigma^2 \left(\sqrt{\frac{\log{d}}{n}} + \frac{\log{d}}{n} \right) + \frac{b^2}{n} + \frac{2b}{\sqrt{n}} \cdot \sqrt{C \sigma^2 \left(\sqrt{\frac{\log{d}}{n}} + \frac{\log{d}}{n} \right)  + \max_{i\in [d]} \Sigma_{ii} }.  
		\end{align*}
        }
       \end{prf}
 
   Next, we establish the following lemma, which transforms the upper bound of $\infnorm{E} / n$ studied in \cref{lem:inf_norm_bound_of_E} into a user-friendly bound:

   \begin{lemma}
       \label{lem:upper_bound_f}
       Define $f:=f(C, \Sigma, \sigma, d, n)$ to be the RHS of \eqref{eqn:inf_norm_bound_of_E}. 
       Assume that $\lambda_1 - \lambda_2 > 0$.
       There exists an absolute constant $C^* > 0$ such that when
{\setlength{\abovedisplayskip}{2pt}
  \setlength{\belowdisplayskip}{0pt}
       \begin{align*}
           n\ge n^*:= \max\left\{C^* \cdot \left[\frac{k^2 \sigma^4\log{d} + b^2 k^2 \left(\sigma^2 + \max \Sigma_{ii} \right)}{(\lambda_1 - \lambda_2)^2 a^4} + \frac{kb^2}{(\lambda_1 - \lambda_2) a^2} \right], \log{d} \right\},
        \end{align*}
        }
        one has $f \le (\lambda_1 - \lambda_2) \cdot a^2 / (8k)$ with probability at least $1 - d^{-10}$.
        Moreover, if for $l\ge 1$, $n\ge l \cdot n^*$, then $f\le (\lambda_1 - \lambda_2) \cdot a^2 / (8\sqrt{l}k)$ with probability at least $1 - d^{-10}$.
   \end{lemma}

   \begin{prf}
       Since $n\ge \log{d}$, one has $1\ge \sqrt{\frac{\log{d}}{n}} \ge \frac{\log{d}}{n}$. 
       By the fact that $\sqrt{x + y} \le \sqrt{x} + \sqrt{y}$ for non-negative scalars $x,y$, it suffices to show that
{\setlength{\abovedisplayskip}{2pt}
  \setlength{\belowdisplayskip}{0pt}
       \begin{align}
       \label{eqn:upper_bound_of_f}
          2 C \sigma^2 \sqrt{\frac{\log{d}}{n}} + \frac{b^2}{n} + \frac{2b}{\sqrt{n}} \cdot \left( \sqrt{2 C \sigma^2 } + \sqrt{\max_{i\in [d]} \Sigma_{ii}} \right) \le \frac{\lambda_1 - \lambda_2}{8k} \cdot a^2.
       \end{align}
       }
       It is then clear that for some large $C^* > 0$, \eqref{eqn:upper_bound_of_f} holds true.

       Suppose, in addition, $n\ge l\cdot n^*$, then by definition of $f$, it is also clear that 
       {\setlength{\abovedisplayskip}{2pt}
  \setlength{\belowdisplayskip}{0pt}
       \begin{align*}
           2 C \sigma^2 \sqrt{\frac{\log{d}}{n}} + \frac{b^2}{n} + \frac{2b}{\sqrt{n}} \cdot \left( \sqrt{2 C \sigma^2 } + \sqrt{\max_{i\in [d]} \Sigma_{ii}} \right) \le \frac{\lambda_1 - \lambda_2}{8\sqrt{l}k} \cdot a^2.
       \end{align*}
       }
        \end{prf}
	
    Combining everything we have developed so far, we are ready to prove \cref{prop:robust random}:
    
    \begin{prfc}[of \cref{prop:robust random}]
		Define $f(C, \Sigma, \sigma, d, n)$ to be the RHS of \eqref{eqn:inf_norm_bound_of_E}.
		By \cref{prop:robust deter} and \cref{lem:inf_norm_bound_of_E}, we obtain that
        {\setlength{\abovedisplayskip}{2pt}
  \setlength{\belowdisplayskip}{0pt}
		\begin{align*}
			  \infnorm{W^* - vv^\top}
	            &\le \|W^* - v v^\top\|_F 
	            \le \frac{2k}{\lambda_1 - \lambda_2} \cdot \frac{1}{n} \infnorm{E} + \frac{1}{\sqrt{n}} \sqrt{\frac{2k}{\lambda_1 - \lambda_2} \cdot \frac{1}{n}\infnorm{E}}\\
				&\le \frac{2k f(C, \Sigma, \sigma, d, n)}{\lambda_1 - \lambda_2}  + \frac{1}{\sqrt{n}} \sqrt{\frac{2k f(C, \Sigma, \sigma, d, n)}{\lambda_1 - \lambda_2}}
		\end{align*}
	        }
		holds with probability at least $1 - d^{-10}$.

        Next, we show that there exists an absolute constant $C^* > 0$ such that when 
        {\setlength{\abovedisplayskip}{2pt}
  \setlength{\belowdisplayskip}{0pt}
        \begin{align*}
           n\ge n^*:= \max\left\{C^* \cdot \left[\frac{k^2 \sigma^4\log{d} + b^2 k^2 \left(\sigma^2 + \max \Sigma_{ii} \right)}{(\lambda_1 - \lambda_2)^2 a^4} + \frac{kb^2}{(\lambda_1 - \lambda_2) a^2} \right], \frac{4}{a^2}, \log{d} \right\},
        \end{align*}
        }
        we have that
        {\setlength{\abovedisplayskip}{2pt}
  \setlength{\belowdisplayskip}{0pt}
        \begin{align*}
            \frac{2k f(C, \Sigma, \sigma, d, n)}{\lambda_1 - \lambda_2}  + \frac{1}{\sqrt{n}} \sqrt{\frac{2k f(C, \Sigma, \sigma, d, n)}{\lambda_1 - \lambda_2}} \le \frac{a^2}{2}.
        \end{align*}
        }
        Note that the above inequality is implied by the fact that
        {\setlength{\abovedisplayskip}{2pt}
  \setlength{\belowdisplayskip}{0pt}
        \begin{align*}
            \sqrt{\frac{2k f(C, \Sigma, \sigma, d, n)}{\lambda_1 - \lambda_2}} \le \frac{-\frac{1}{\sqrt{n}} + \sqrt{\frac{1}{n} + 2a^2}}{2} = \frac{a^2}{\sqrt{\frac{1}{n} + 2a^2} + \frac{1}{\sqrt{n}}}.
        \end{align*}
        }
        Then it is clear that when $n\ge 4 / (a^2)$, one has that $\frac{a^2}{\sqrt{\frac{1}{n} + 2a^2} + \frac{1}{\sqrt{n}}} \ge \frac{a^2}{2a} = \frac{a}{2}$.
        Therefore, it suffices to show that for some $n$ large enough
        {\setlength{\abovedisplayskip}{2pt}
  \setlength{\belowdisplayskip}{0pt}
        \begin{align*}
            \sqrt{\frac{2k f(C, \Sigma, \sigma, d, n)}{\lambda_1 - \lambda_2}} \le \frac{a}{2} \quad & \Longleftrightarrow \quad
            \frac{2k f(C, \Sigma, \sigma, d, n)}{\lambda_1 - \lambda_2} \le \frac{a^2}{4}
            \Longleftrightarrow \quad
            f\le \frac{\lambda_1 - \lambda_2}{8k} \cdot a^2,
        \end{align*}
        }
        which is then implied by \cref{lem:upper_bound_f}.
\end{prfc}

        Finally, we present the proof of \cref{thm:randomized_alg_in_stat_model}.
        The high level idea in the proof is to make use of the fact that $W^*$ is close to $vv^\top$ as developed in \cref{prop:robust random}, and hence there is a simple way to control the quality of the solution we obtained in our algorithm.

     \begin{prfc}[of \cref{thm:randomized_alg_in_stat_model}]
        In the proof, we will constantly use the following fact:
\begin{fact}
	\label{fact:holder}
	Suppose $x\in\R^d$ is a unit $k$-sparse vector, then for a $d\times d$ matrix $E$, one has $\abs{x^\top E x} \le k\infnorm{E}$.
\end{fact}
\cref{fact:holder} can be shown via Holder's inequality and the fact that $\onorm{xx^\top} = \onorm{x}^2 \le k \norm{x}^2 = 1$.

We write $A = n \Sigma + \left(B^\top B - n \Sigma \right) +  \left(M^\top M + M^\top B + B^\top M\right) =: n\Sigma + E$.
It is clear that 
{\setlength{\abovedisplayskip}{2pt}
  \setlength{\belowdisplayskip}{0pt}
\begin{align*}
	\frac{1}{n} (x^*)^\top A x^* 
	&\ge \max_{\substack{\norm{x} = 1\\\znorm{x}\le k}} x^\top \Sigma x - \max_{\substack{\norm{x} = 1\\\znorm{x}\le k}} x^\top \left(\frac{1}{n}E\right) x 
    = \lambda_1 - \max_{\substack{\norm{x} = 1\\\znorm{x}\le k}} x^\top \left(\frac{1}{n}E\right) x
    \ge \lambda_1 - \frac{k}{n} \infnorm{E},
\end{align*}
}
where the first inequality uses the basic proposition of $\max$ function, and the second inequality follows from \cref{fact:holder}. 
Similarly, one can obtain that 
{\setlength{\abovedisplayskip}{2pt}
  \setlength{\belowdisplayskip}{0pt}
\begin{align}
\label{eqn:upper_bound_opt}
	\frac{1}{n} (x^*)^\top A x^* \le \lambda_1 + \frac{k}{n} \infnorm{E}.
\end{align}
}
We denote $\bar x$ the heuristic solution obtained via lines 1 - 2 in \cref{alg:multi_ra}.
By the definition of $\bar x$, and the fact that $\bar x$ shares the same support with $v$ by \cref{prop:robust random} with high probability by~\cref{prop:robust random}, one can see that $\frac{1}{n} \bar x^\top A \bar x		\ge 	\frac{1}{n} v^\top A v
		=  v^\top \Sigma v + \frac{1}{n} v^\top E v
		\ge \lambda_1 - \frac{k}{n}\infnorm{E}$
where we use \cref{fact:holder} in the last inequality.
Summing everything up, we see that
\begin{equation}
    \label{eqn:lower_bound_opt}
\begin{aligned}
	\frac{(x^*)^\top A x^* - \bar x^\top A \bar x}{n} 
	& \le \frac{1}{n} (x^*)^\top A x^*  - \lambda_1 + \frac{k}{n}\infnorm{E}
    \le \frac{1}{n}\tr(AW^*) - \lambda_1 + \frac{k}{n}\infnorm{E}
	\le \frac{2k}{n}\infnorm{E}.
\end{aligned}
\end{equation}
	Define $\epsilon:=\frac{2}{8\sqrt{l} - 1}$.
	Next, we show that
\begin{equation}
    \label{eqn:epsilon}
    \frac{\lambda_1 - \lambda_2}{8\sqrt{l} k}\cdot a^2 \le  \frac{\lambda_1}{k} \cdot \frac{\epsilon}{2 + \epsilon}.
\end{equation}
	In fact, \eqref{eqn:epsilon} is equivalent to $2 (\lambda_1 - \lambda_2) a^2  \le \left[ 8\sqrt{l}\lambda_1 - (\lambda_1 - \lambda_2) a^2 \right] \cdot \epsilon$.
	Since $0 < \lambda_1 - \lambda_2 \le \lambda_1$ and $a\le 1$, it is enough to check that
	\begin{equation*}
	    2\lambda_1 \le (8\sqrt{l}\lambda_1 - \lambda_1)\cdot \epsilon,
	\end{equation*}
	which holds by the definition of $\epsilon$.

	Finally, combining the lower bound on $\bar x^\top A \bar x$ above with \eqref{eqn:upper_bound_opt}, one obtains that with probability at least $1 - d^{-10}$,
	\begin{equation}
	    \label{eqn:upper_bound_on_infnorm_of_E_v2}
		\frac{2k}{n} \infnorm{E} \le \epsilon \left(\lambda_1 - \frac{k}{n}\infnorm{E}\right) \le \epsilon \cdot \frac{1}{n} \cdot \bar x^\top A \bar x, 
	\end{equation}
	where the first inequality is implied by \cref{lem:inf_norm_bound_of_E}, \cref{lem:upper_bound_f}, and \eqref{eqn:epsilon}: $\frac{1}{n}\infnorm{E}\le \frac{\lambda_1 - \lambda_2}{8\sqrt{l} k}\cdot a^2 \le  \frac{\lambda_1}{k} \cdot \frac{\epsilon}{2 + \epsilon}$.
		The desired approximation ratio is then given by \eqref{eqn:lower_bound_opt} and \eqref{eqn:upper_bound_on_infnorm_of_E_v2}, since these inequalities imply
		\begin{equation*}
		(x^*)^\top A x^* - \bar x^\top A \bar x \le \epsilon \, \bar x^\top A \bar x,
		\end{equation*}
		and therefore $(x^*)^\top A x^* \le (1+\epsilon)\bar x^\top A \bar x$.
\end{prfc}
\end{APPENDICES}
}{%
\appendix

}%
}

\newcommand{\IJOOPrintBibliography}{%
\ifthenelse{\boolean{IJOO}}{%
     \bibliographystyle{informs2014}
     \bibliography{biblio}
}{%
\bibliographystyle{plain}
\bibliography{biblio}
}%
}

\begin{document}

\maketitle

\ifthenelse{\boolean{IJOO}}
{}
{
\begin{abstract}
\bstrct
\end{abstract}
\bigskip

\noindent
\hangindent=2cm
\emph{Key words:}
\kywrds
}


\ifIJOOincludeMain

\newcommand{\note}[1]{\color{magenta} Note: #1 \color{black}}

\section{Introduction}
	\label{sec:intro spca}
	
	The Principal Component Analysis (PCA) problem involves finding a linear combination of $d$ features that captures the maximum possible variance in a given $d\times d$ data matrix $A$. 
	Formally, the problem is defined as
    {
        \setlength{\abovedisplayskip}{0pt}
        \setlength{\belowdisplayskip}{0pt}%
	\begin{align}
        \max \ x^\top A x \quad \text{s.t. } \norm{x} = 1. \tag{PCA}  \label{prob PCA}
	\end{align}
    }
	\ref{prob PCA} is a widely used statistical technique for reducing the dimensionality of large datasets, and it has been successfully applied to a broad range of topics, including neuroscience, meteorology, psychology, genetics, finance, and pattern recognition.
	For a comprehensive overview of the applications of \ref{prob PCA}, we refer interested readers to \cite{jolliffe2002principal}.

	Despite its usefulness, the interpretation of a solution to PCA is limited since the principal component (PC) is often a linear combination of all $d$ features. 
	To address this issue, the Sparse Principal Component Analysis problem (SPCA) is introduced.
	SPCA aims to find a \emph{sparse} linear combination of features while capturing the maximum variance.
	SPCA is formally defined as:
    {\setlength{\abovedisplayskip}{2pt}
        \setlength{\belowdisplayskip}{2pt}%
	\begin{align}
        \max \ x^\top A x \quad \text{s.t. } \norm{x} = 1, \ \ \znorm{x}\le k. \tag{SPCA}  \label{prob SPCA}
	\end{align}}
	Here $k$ is the \emph{sparsity constant,} a positive integer that sets an upper bound on the number of non-zero entries in the $d$-dimensional vector $x$.
	Throughout the paper, we take $A$ to be symmetric. 
	In the statistical settings that motivate \ref{prob SPCA}, $A$ is typically positive semidefinite, but some of our approximation guarantees extend to more general symmetric matrices with non-negative diagonal entries. 
	Compared with \ref{prob PCA}, using \ref{prob SPCA} for dimensionality reduction yields more interpretable components, lowers downstream memory and computational costs, and helps prevent overfitting. 
	\ref{prob SPCA} has a wide range of real-world applications, such as identifying influential single-nucleotide polymorphisms in genetics~\cite{lee2012sparse}, selecting informative object features in computer vision~\cite{naikal2011informative}, and organizing a large corpus of text data in data science~\cite{zhang2011adaptive}. 
	
	\ref{prob SPCA} is an NP-hard problem in general~\cite{magdon2017np}, making it computationally challenging in practice.
	In fact, it is NP-hard to approximate \ref{prob SPCA} within a multiplicative ratio of $(1+\epsilon)$ for some constant $\epsilon>0$~\cite{chan2015worst}. 
	Despite the hardness of the problem, many polynomial-time approximation algorithms have been proposed to tackle \ref{prob SPCA}. 
	For instance, \cite{papailiopoulos2013sparse} areccelerated \ref{prob SPCA} by replacing the input matrix with its rank-$m$ approximation formed from the top $m$ eigenpairs, resulting in an algorithm with a running time $\mo(d^{m+1}\log d)$ and an approximation ratio $1 / (1 - \delta_m)$, with $\delta_m \le \lambda_{m + 1} \cdot \min\{d  / (k \lambda_1), 1 / \max_{i \in [d]} A_{ii} \}$. 
	This algorithm has the advantage that if the eigenvalues $\lambda_1\ge\lambda_2\ge\cdots\ge\lambda_d$ of $A$ satisfy an exponential decay, i.e., $\lambda_{i+1}/\lambda_i\le c\in(0,1)$, then $\delta_m$ decreases exponentially as $m$ increases. 
    \cite{li2020exact} introduced two polynomial-time $k$-approximation algorithms: the Greedy algorithm and the Local Search algorithm.
    \cite{chan2016approximability} introduced an $\epsilon$-additive approximation algorithm with a runtime of $\mo(d^{\text{poly}(1/\epsilon)})$ based on the previous work~\cite{asteris2015sparse}.
	Moreover, \cite{chan2016approximability} also proposed a simple $\min\{\sqrt{k},d^{1/3}\}$-approximation algorithm for \ref{prob SPCA}, which is, to the best of our knowledge, the best known approximation ratio for \ref{prob SPCA} algorithms with polynomial runtime.

	\textbf{Our contributions.}
	In this paper,  we propose a randomized algorithm based on the \emph{basic} semidefinite programming (SDP) relaxation to \ref{prob SPCA}, 
    {
     \setlength{\abovedisplayskip}{2pt}
    \setlength{\belowdisplayskip}{2pt}%
	\begin{align}
		\label{prob SDP SPCA} \tag{SPCA-SDP}
		\max \ \tr(A W) \quad \text{s.t. } \tr(W) = 1, \ \onorm{W}\le k, \ W \succeq 0.
	\end{align}
    }
    Here $\onorm{W}:=\sum_{i,j}|W_{ij}|$ denotes the entrywise $1$-norm of $W$.
    At a high level, given an (approximate) optimal solution $W^*$ of \ref{prob SDP SPCA}, our algorithm constructs two types of feasible solutions: a deterministic solution obtained from the largest diagonal entries of $W^*$, and a number of randomized solutions generated from $W^*$. 
	These two components have complementary merits - the randomized solutions give our general approximation guarantees, while the deterministic solution becomes especially effective when the input matrix $A$ contains a certain hidden sparse structure. 
	The algorithm finally returns the best solution among them, so the same SDP-based procedure can succeed for different reasons on different classes of instances.
    This algorithm is not only efficient, but also improves upon the best known polynomial-time benchmark of $\min\{\sqrt{k},d^{1/3}\}$ across a variety of practically motivated instances.

    {\it Efficient SDP-based Algorithm with Provable Guarantees.} 
	Our algorithm transforms an (approximate) optimal solution $W^*$ of \ref{prob SDP SPCA} into a randomized solution for \ref{prob SPCA}, in time $\mo(d\log{d})$.
	Despite its efficiency, by repeating this randomized rounding step independently $\Omega(d/k)$ times, our algorithm achieves a $k$-approximation with high probability, matching the guarantees of the Greedy algorithm and the Local Search algorithm in \cite{li2020exact}.
    The average approximation ratio of this randomized component can be further upper bounded by $\mo(\log{d})$ under the mild condition $k=\Omega(\log(d/k))$ and a technical assumption related to sum of square roots of diagonal entries of $W^*$.
    We note that in the regime where $k = \Omega(\log^2{d})$, and where the technical assumption is true, our algorithm admits an average approximation error strictly better than the best-known polynomial-time guarantee $\min\{\sqrt{k},d^{1/3}\}$.
    We further show that this technical assumption is true if $W^*$ is of a fixed rank, or $W^*$ admits exponentially decaying eigenvalues.
    We further identify two classes of instances that guarantee a rank-one optimal solution, and hence satisfy the technical assumption:  
    (i) Rank-one input matrices whose non-zero entries are bounded below by a certain value; 
    (ii) General input matrices whose support block has a sufficiently separated leading eigenvalue, for which a certain resolvent ratio reaches the target level $\sqrt{k}$ with sufficient spectral margin, and whose off-support entries are sufficiently small.
    To the best of our knowledge, we provide the first deterministic classes of instances in which \ref{prob SDP SPCA} admits rank-one optimal solutions. 
    These structural results are important because they identify concrete classes in which the basic SDP relaxation is already highly informative, with a rank-one optimal solution that captures the relevant sparse structure of the instance.
    We also perform extensive numerical experiments, demonstrating the effectiveness of our proposed algorithm in real-world datasets. 
    Oftentimes, our algorithm can find a solution which is as good as the best solution other existing methods studied in \cite{chan2016approximability,papailiopoulos2013sparse,li2020exact,berk2019certifiably,dey2022solving} can find.
    In terms of the runtime for \ref{prob SDP SPCA}, although general‐purpose SDP solvers scale poorly, we find that our GPU implementation, based on the approximation algorithm in~\cite{yurtsever2019conditional}, proves remarkably efficient: it finds $W^*$ on $d=2000$ instances of \ref{prob SDP SPCA} in under six seconds.
    
    {\it Adversarial-Robust Recovery Guarantees.} We show that \ref{prob SDP SPCA} is robust to adversarial perturbations in a general covariance model. 
    Specifically, we consider a scenario where the input matrix $A$ is under adversarial attack, i.e., $A = (B + M)^\top (B + M)$, where $B$ represents the data matrix that has i.i.d.~rows sampled from a covariance model having a sparse spike, and $M$ represents a bounded adversarial perturbation. 
    We show that \ref{prob SDP SPCA} still yields an optimal solution close to the sparse spike when the sample size is sufficiently large. 
    This generalizes the findings of \cite{d2020sparse} and provides further insight into the strong computational performance of \ref{prob SDP SPCA}. 
	In this model, the near-one guarantee comes primarily from the deterministic solution in our pipeline rather than from the randomized solutions: once $W^*$ is close to the sparse spike, the solution formed from the largest diagonal entries of $W^*$ is already nearly optimal. 
	Thus, even when the technical assumption used in our randomized analysis is not met, \ref{prob SDP SPCA} and our overall algorithm can still perform exceptionally well across diverse inputs.

	\textbf{Organization of this paper.}
    In \cref{sec:related_work_spca}, we provide an overview of related work.
	In \cref{sec:ra for spca}, we introduce our SDP-based randomized algorithm for solving \ref{prob SPCA} and provide theoretical results on approximation guarantees, focusing on the guarantees driven by the randomized solutions and on structural conditions that make the randomized solutions strong.
	In \cref{sec:adversarial pert}, we define a general statistical model that considers adversarial perturbation, demonstrate the robustness of \ref{prob SDP SPCA} against such perturbations, and show that in this model the deterministic solution achieves a near-optimal approximation bound.
	In \cref{sec:numerical tests spca}, we present numerical experiments conducted on various real-world datasets to evaluate the performance of our algorithm and compare with other algorithms.
	We defer some proofs in \cref{sec:ra for spca,sec:adversarial pert} to Appendices~\ref{sec:proof of ra} and \ref{sec:proof of adver}.
    For the remainder of the section, we introduce the notation used in the paper.

	\textbf{Notation.} 
    \textbf{Sets, vectors, and matrices:}
	For any positive integer $d$, we define $[d]:=\{1, 2, \ldots, d\}$.
	Let $x$ be a $d$-vector.
	The \emph{support} of $x$ is the set $\supp(x) := \{i\in[d]:x_i\ne0\}$.
    Given an index set $\mathcal{I}\subseteq [d]$, denote by $x_{\mathcal{I}}$ the sub-vector of $x$ indexed by $\mathcal{I}$, and we write $x_{i}:=x_{\{i\}}$.
	For $1\le p \le \infty$, we denote the \emph{$p$-norm} of $x$ by $\|x\|_p$.
	The \emph{$0$-(pseudo)norm} of $x$ is
	$\znorm{x}:=|\supp(x)|.$
	We say that $x$ is \emph{$k$-sparse} if $\znorm{x}\le k$.
	Let $M$ be a $m\times n$ matrix.
	Given two index sets $\mathcal{I}\subseteq [m]$, $\mathcal{J}\subseteq [n]$, we denote by $M_{\mathcal{I}, \mathcal{J}}$ the submatrix of $M$ consisting of the entries in rows $\mathcal{I}$ and columns $\mathcal{J}$.  
	Let $X$ be an $m \times m$ symmetric positive semidefinite matrix, i.e., $X \succeq 0$. We denote by $\sqrt{X}$ the \emph{matrix square root of $X$}, i.e., $\sqrt{X}=(\sqrt{X})^\top\succeq 0$ and $X = \sqrt{X}\sqrt{X}$.
	For $1\le p$, $q\le \infty$, the \emph{$p$-to-$q$ norm} of $M$ is defined as $\|M\|_{p\rightarrow q}:=\max_{\|x\|_{p}=1} \|Mx\|_q.$
	The \emph{2-norm} of $M$ is defined by $\|M\|_2=\|M\|_{2\rightarrow2}$.
	The \emph{1-norm} of $M$ is defined by $\onorm{M} = \sum_{i, j}|M_{ij}|$.
	The \emph{infinity norm} of $M$ is defined by $\maxnorm{M}:=\max_{i,j} |M_{ij}|$. 
	The \emph{Frobenius norm} of $M$ is defined as $\fnorm{M}:=\sqrt{\sum_{i,j} |M_{ij}|^2}$. \\
	\textbf{Approximation ratio and $\epsilon$-approximate solution:}
	Let $w^*$ be an optimal solution to a maximization problem $\mathcal{P}$ with objective function $f$ and input $D$. 
	We say a (randomized) algorithm $\mathcal{A}$ is an approximation algorithm to $\mathcal{P}$ with an \emph{approximation ratio} $r$, if $\mathcal{A}$ can output a random solution $\bar w$ with input $D$ such that $\me f(\bar w) \ge 1/ r \cdot f(w^*)$.
	Sometimes we will also say that $\mathcal{A}$ is an \emph{$r$-approximation algorithm} for brevity. 
	We say a solution $\tilde w$ is an \emph{$\epsilon$-approximate solution} to $\mathcal{P}$ if $\tilde w$ is feasible to $\mathcal{P}$ such that $f(\tilde w) \ge f(w^*) - \epsilon$.
    
\section{Related work}
\label{sec:related_work_spca}

In this section, we discuss the literature related to our work.
First, we discuss results related to the basic SDP relaxation, \ref{prob SDP SPCA}.
It is known that SDP can be solved in polynomial time up to an arbitrary accuracy, by means of the ellipsoid algorithm and interior point methods~\cite{vandenberghe1996semidefinite,LauRen05}.
    The basic SDP relaxation \ref{prob SDP SPCA} was initially proposed in~\cite{d2004direct} and has been extensively researched since then.  
	The literature includes studies on its performance under various statistical models and its approximability.
	The statistical performance of \ref{prob SDP SPCA} has been thoroughly investigated, with the assumption that $A = B^\top B$ and $B$ being an $n\times d$ matrix. 
	For example, \cite{AmiWai08} demonstrated that the \emph{sparse spike}, which is the sparse maximal eigenvector, can be recovered in a particular covariance model, known as the Wishart spiked model, when the number of samples $n$ is above the threshold $\Omega(k\log{d})$. 
	Then, \cite{krauthgamer2015semidefinite} showed that \ref{prob SDP SPCA} is unable to recover the sparse spike if $k = \Omega(\sqrt{n})$ in the model discussed in \cite{AmiWai08}. 
	In a more general spiked covariance model, \cite{wang2016statistical} showed that \ref{prob SDP SPCA} can recover the sparse spike but at a slightly higher sample complexity $\Omega(k^2\log{d})$. 
	Additionally, \cite{d2020sparse} demonstrated that \ref{prob SDP SPCA} is robust to adversarial perturbations in the Wishart spiked model.
	Moreover, a line of work~\cite{berthet2013computational,berthet2013optimal} investigated the information theoretical limits of \ref{prob SDP SPCA} recovering the sparse spike in certain covariance models.
	Regarding approximation results, an approximation algorithm based on \ref{prob SDP SPCA} was developed in~\cite{chowdhury2020approximation}. 
	It is worth noting that authors of \cite{chowdhury2020approximation} acknowledged that their theoretical guarantees may not be indicative of the outstanding practical performance of their algorithm. 
	Nevertheless, they provided compelling empirical evidence of its efficacy by showcasing impressive computational results on diverse real-world datasets.
	A worst-case approximation bound of \ref{prob SDP SPCA} was studied in \cite{chan2016approximability}, revealing there exists an instance that results in an approximation ratio that is quasi-quasi-polynomial in $d$.

    Then, we discuss efficient algorithms for \ref{prob SPCA} in the literature.
	To the best of our knowledge, there are currently five other main categories of methods for finding (approximate) solutions to \ref{prob SPCA}, except via its basic SDP relaxation.
	Firstly, various existing methods solve \ref{prob SPCA} by relaxing the sparsity constraint with a convex constraint. 
	These methods include providing practical algorithms to maximize the objective value in an $\ell_1$ ball~\cite{dey2017sparse}, or by solving stronger SDP relaxations~\cite{kim2019convexification}. 
	Secondly, methods based on integer programming have been developed to solve \ref{prob SPCA} exactly or approximately, including using integer programs to obtain dual bounds, for either single sparse PC~\cite{dey2022using} or multiple but row-sparse PCs~\cite{dey2022solving}, solving mixed-integer SDPs or mixed-integer linear programs~\cite{li2020exact}, using branch-and-bound algorithm to obtain certifiable (near) optimality~\cite{berk2019certifiably,bertsimas2022solving}, and combining integer programs with geometric approach to obtain multiple sparse PCs~\cite{bertsimas2022sparse}.
	Thirdly, polynomial-time algorithms for a fixed rank input matrix $A$ are proposed, either for a single sparse PC~\cite{papailiopoulos2013sparse}, or for multiple but row-sparse PCs~\cite{del2022sparse}.
	It should be noted that the complexity of these algorithms are oftentimes exponential in $\rk(A)$.
	Fourthly, polynomial running-time approximation algorithms for \ref{prob SPCA} have been developed, including finding a low rank approximation of $A$ and then solving \ref{prob SPCA} exactly~\cite{papailiopoulos2013sparse}, deriving an approximation algorithm via basis truncation~\cite{chan2016approximability}, and via basic SDP relaxation~\cite{chowdhury2020approximation}.
	The fifth category includes methods for solving~\ref{prob SPCA} in certain statistical models via different approaches, including covariance thresholding~\cite{deshpande2014sparse} and methods that combine diagonal thresholding with exhaustive search over small index sets~\cite{ding2023subexponential}.
    A recent work~\cite{delpia2025efficient} introduces a plug-and-play framework to provide speedup to these existing algorithms in the categories by first finding block-diagonal approximation to the input matrix $A$ and then solving \ref{prob SPCA} sub-problems inside each block.
	For a more comprehensive review of the literature in each of the categories mentioned above, interested readers are referred to the cited papers.
	
	Note that some of the aforementioned work have established approximation results using non-convex optimization methods, which may necessitate an exponential runtime. 
    The non-convexity comes from the fact that the authors solve a maximization problem of a convex objective.
	For instance, the authors of \cite{dey2022using} have demonstrated that by relaxing \ref{prob SPCA} within an $\ell_1$ ball, an upper bound for \ref{prob SPCA} with constant approximation ratio can be achieved. 
	In \cite{dey2022solving}, the authors have extended these findings to finding $m$ sparse principal components with a global support, with an approximation ratio $\mo(\sqrt{\log{m}})$.
\section{A randomized algorithm for SPCA}
	\label{sec:ra for spca}

	In this section, we present our randomized approximation algorithm for \ref{prob SPCA}. 
	For brevity, we denote an $\epsilon$-approximate optimal solution to \ref{prob SDP SPCA} as $W^*$ and an optimal solution to \ref{prob SPCA} as $x^*$, for the rest of this paper.

    We begin by presenting the motivation. 
    Although \ref{prob SDP SPCA} has nice properties, such as the result in \cite{li2020exact}, where the authors demonstrate that the objective value of an optimal solution to \ref{prob SDP SPCA} is at most $k$ times that of \ref{prob SPCA}, there are several limitations that hinder the practical application of \ref{prob SDP SPCA}. 
    For instance, as pointed out in \cite{deshpande2014sparse}, that there are also some theoretical limitations of \ref{prob SDP SPCA}. 
	Specifically, (i) $W^*$ is not guaranteed to be a rank-one matrix in general; and (ii) in some cases where $W^*$ is rank-one, denoted as $W^* = v^* (v^*)^\top$, but oftentimes the zero-(pseudo)norm of $v$, $\znorm{v}$, is larger than $k$. 
	This raises a natural question: is there a way to transform $W^*$ into a feasible solution for \ref{prob SPCA} with high quality?
	
	In \cite{chowdhury2020approximation}, the authors partially address this question by providing a vector $z$, obtained by finding the best rank-one approximation $uu^\top$ to $W^*$, keeping the $\mathcal{O}(k^2/\epsilon^2)$ largest components (in absolute value) in $u$, and setting the other entries to zero. 
	They obtain that $z^\top A z \ge 1/\alpha \cdot (x^*)^\top A x^* - \epsilon$, where $\alpha \ge 1$ is the ratio $\tr(AW^*) / u^\top A u$. However, this algorithm has two major issues: (a) $z$ is generally not a $k$-sparse vector and $\znorm{z}$ could be much larger than $k$; and (b) there is no clear theoretical bound on $\alpha$, making it difficult for users to predict the worst-case quality of $z$.
	
	In this section, we present \cref{alg:randomized spca simple} as an approximation algorithm for \ref{prob SDP SPCA} with the aim of obtaining a high-quality $k$-sparse vector from $W^*$ and addressing the issues discussed above. 
	The main idea behind \cref{alg:randomized spca simple} is to treat the diagonal entries in $W^*$ and $A$ as probability masses that determine whether or not to include the corresponding index in the support of a vector $x$. 
	For a positive semidefinite matrix $W\in \R^{d\times d}$, define its sum of square roots by $\textup{SSR}(W):=\sum_{i=1}^d \sqrt{W_{ii}}$. When the underlying matrix is clear from context, we simply write $\textup{SSR}$; in particular, in statements involving $W^*$ we use $\textup{SSR}:=\textup{SSR}(W^*)$. 
	
\begin{algorithm}
	\caption{Randomized Algorithm for \ref{prob SPCA}}
	\label{alg:randomized spca simple}
	\begin{algorithmic}[1] 
		\REQUIRE A matrix $A\in \R^{d\times d}$, a positive semidefinite matrix $W\in \R^{d\times d}$, and a positive integer $k$
		\ENSURE A vector $ z \in \R^{d}$, such that $\norm{z} = 1$ and $\znorm{z} \le k$ with high probability
		\FOR{$i = 1$ {\bfseries to} $d$} 
        \STATE $a_i\gets \sqrt{W_{ii}}$
        \ENDFOR
		\FOR{$i = 1$ {\bfseries to} $d$}
		\STATE $p_i \gets \min\{1, 2 / 3 \cdot k a_i / \textup{SSR}(W) + 1 / 12 \cdot kA_{ii} / \tr(A)\}$
		\STATE Sample independently $\epsilon_i \gets 1$ with probability $p_i$, and $\epsilon_i \gets 0$ with probability $1 - p_i$
		\ENDFOR
  
        \STATE $S \gets \{i\in[d]: \epsilon_i = 1\}$, $z\gets \text{zero vector in $\R^d$}$

        \IF{$A\succeq 0$ \AND $|S| < k$}
        \STATE $S \gets S \cup T$, with $T\cap S = \emptyset$ and $|T| = k - |S|$
        \ENDIF
        
        \STATE $z_S\gets \argmax_{\norm{y} = 1} y^\top A_{S,S} y$
		\STATE \textbf{return} {$z$}
	\end{algorithmic}
\end{algorithm}
\vspace{-5pt}
	    In \cref{alg:randomized spca simple}, lines 7--9 are included only to improve the practical output.
	    The choice of $T$ on line 8 does not affect the theoretical bounds in the next section.
	    In our experiments in \cref{sec:numerical tests spca}, we set $T$ to be the indices in $[d]\backslash S$ with the largest values of $a_i$ (equivalently, of $W_{ii}$), which can be found in time $\mo(d\log{d})$.
	    Except for line 8, \cref{alg:randomized spca simple} requires runtime at most $\mo(d + k^2\log{k})$ (assuming line 9 is solved through randomized Lanczos method~\cite{royer2020newton}).

    In practice, one can call \cref{alg:randomized spca simple} several times, with the intention of obtaining a better solution. 
    The operational procedures are presented in \cref{alg:multi_ra}.
    Approximation guarantee for these algorithms will be provided in the subsequent sections.
    
    \begin{algorithm}
	\caption{Multi-run Randomized Algorithm for SPCA}
	\label{alg:multi_ra}
	\begin{algorithmic}[1] 
		\REQUIRE A matrix $A\in \R^{d\times d}$, a positive semidefinite matrix $W\in \R^{d\times d}$, a positive integer $k\le d$, and a positive integer $N$ denoting the number of independent calls to \cref{alg:randomized spca simple}
		\ENSURE A unit vector $ z \in \R^{d}$, such that $\znorm{z} \le k$
      \STATE $S_0\gets$ the set of indices in $[d]$ that corresponds to the $k$ largest diagonal entries in $W$ 
      \STATE $z_0\gets \argmax_{\norm{y} = 1} y^\top A_{S_0,S_0} y$
		\FOR{$i = 1$ {\bfseries to} $N$}
            \STATE Obtain $z_i$ using \cref{alg:randomized spca simple} with input $(A, W, k)$
        \ENDFOR
		
		\STATE \textbf{return} the best feasible solution to \ref{prob SPCA} among $\{z_i\}_{i = 0}^N$
	\end{algorithmic}
	\end{algorithm}
	
    Note that lines 1--2 in \cref{alg:multi_ra} define a deterministic solution, which already guarantees that \cref{alg:multi_ra} outputs at least one feasible solution to \ref{prob SPCA}. 
	Lines 3--5 then generate randomized solutions from $W^*$. 
	The general approximation results developed in this section are primarily driven by these randomized solutions, whereas the deterministic solution becomes especially powerful in the statistically structured regimes studied later in \cref{sec:adversarial pert}.

    \subsection{Approximation guarantees}
    \label{sec:app_guarantee}
    
    In this section, we establish approximation bounds for \cref{alg:randomized spca simple,alg:multi_ra}. 
	To streamline the presentation, we defer some of the longer proofs to Appendix~\ref{sec:proof of ra}.

	We begin with the guarantees that come from the randomized solutions in \cref{alg:multi_ra}.
    We first show that, if $N = \Omega(d/k)$, \cref{alg:multi_ra} is a $k$-approximation algorithm with high probability.
    \begin{theorem}
	\label{thm:spca rand alg v2}
	Let $x^*$ be an optimal solution to \ref{prob SPCA} with input pair $(A, k)$, where we assume $A\in \R^{d\times d}$ is positive semidefinite, and $k$ is a positive integer such that $k\le d$.
	Let $W^*$ be an $\epsilon$-approximate optimal solution to \ref{prob SDP SPCA} with input pair $(A, k)$.
    Let $z$ be the output of \cref{alg:multi_ra} with input tuple $(A, W^*, k, N)$.
	
	    Then, \cref{alg:multi_ra} produces a feasible solution $z$ to \ref{prob SPCA} such that
		$z^\top A z \ge (x^*)^\top A x^* / k$ with probability at least $1 - \exp\left\{-kN / (48d)\right\}$.
	\end{theorem}
    Then, we show that the approximation ratio of \cref{alg:randomized spca simple} is also controlled by the \emph{sum of square roots} parameter $\textup{SSR}:=\textup{SSR}(W^*)$, through the factor $\mo(\textup{SSR}^2/k)$.
    Due to the intricate technical details, we present an informal statement of the result as follows. 
    A formal statement can be found in Appendix~\ref{sec:proof of app guarantee}. 
    Additionally, the formal theorem there is more comprehensive: it shows that \cref{alg:randomized spca simple} remains effective even when $A$ is indefinite, since it requires only that $A$ have non-negative diagonal entries rather than be positive semidefinite.
	
	\begin{theorem}[Informal version of \cref{thm:spca rand alg formal}]
		\label{thm:spca rand alg}
		Let $x^*$ be an optimal solution to \ref{prob SPCA} with input pair $(A, k)$, where we assume $A\in \R^{d\times d}$ is positive semidefinite with $\norm{A} = 1$, and $k$ is a positive integer such that $k\le d$.
		Let $W^*$ be an $\epsilon$-approximate optimal solution to \ref{prob SDP SPCA} with input pair $(A, k)$.
		Denote $z$ to be the output of \cref{alg:randomized spca simple} with input tuple $(A, W^*, k)$.
        Then, there exists a high-probability random event $\mathcal{R}\subseteq \{\znorm{z}\le k\}$ such that $\mp(\mathcal{R})\ge 1 - \exp\{-ck\} - 2d^{-3}$ for an absolute constant $c > 0$, and such that when $ck\ge 3\log{(d/k)} + \log\log{d}$, one has
        {\setlength{\abovedisplayskip}{3pt}
  \setlength{\belowdisplayskip}{3pt}%
        \begin{equation*}
                C \log{d} \cdot \Big[1 + \frac{9\textup{SSR}^2}{4k} \Big] \cdot  \me [ z^\top A z \vert \mathcal{R} ] \ge \Big[1  -  \mathcal{O}(\frac{1}{\log{d}} )\Big] (x^*)^\top A x^* - \epsilon,
        \end{equation*}
        }
        for some absolute constant $C > 0$.
	\end{theorem}

	\begin{remark}
        \label{rmk:tech_cond}
		In this remark, we discuss the approximation ratio of \cref{alg:randomized spca simple} in \cref{thm:spca rand alg}.
		On one hand, in \cref{thm:spca rand alg}, we obtain a worst-case multiplicative ratio $\mo(d \log{d} / k)$ due to the fact that $\tr(W^*) = 1$ and Cauchy-Schwarz inequality.
		On the other hand, it is worth noting that when
        {\setlength{\abovedisplayskip}{3pt}
  \setlength{\belowdisplayskip}{3pt}%
		\begin{equation}
			\label{eqn:spca tech cond}
			\textup{SSR} \le c_0 \cdot \sqrt{k}
		\end{equation}
        }
		for some absolute constant $c_0 > 0$, \cref{alg:randomized spca simple} can obtain a  multiplicative ratio $\mo(\log{d})$.
        When $k = \Omega(\log^2{d})$, implying $k\ge 3\log(d/k) + \log\log{d}$, this $\mo(\log{d})$ guarantee strictly surpasses the $\min\{\sqrt{k},d^{1/3}\}$-approximation of \cite{chan2016approximability}, which, to our knowledge, is the smallest ratio previously known for any polynomial-time SPCA algorithm.
        We note that, while \eqref{eqn:spca tech cond} might not always hold, it is easily checkable once \ref{prob SDP SPCA} is solved, which could be done in polynomial time.
        In \cref{sec:tech_cond}, we provide further discussions about assumptions on $W^*$ such that \eqref{eqn:spca tech cond} holds, and in \cref{sec:rank_one}, we provide classes of instances where \eqref{eqn:spca tech cond} holds for $c_0 = 1$.
		Furthermore, as we will see in \cref{sec:numerical tests spca}, \eqref{eqn:spca tech cond} oftentimes holds in our numerical tests (in fact, $c_0 \le 2.21$ for 80\% of the instances).
		The choices of $\mathcal{R}$, $c$, and $C$ are detailed in Appendix~\ref{sec:proof of app guarantee}.
	\end{remark}
	
	\begin{remark}
		\label{rmk:universality}
        In this remark, we point out that \cref{thm:spca rand alg} in fact holds true for any $W^*\succeq 0$ such that $\tr(W^*) = 1$ and $\tr(AW^*)\ge (x^*)^\top A x^* - \epsilon$.
        This implies that our algorithm extends to any SDP relaxation stronger than \ref{prob SDP SPCA}.
        For instance, our rounding scheme applies to the tighter relaxation of \cite{kim2019convexification} and preserves the guarantees of \cref{thm:spca rand alg}.
        We nevertheless focus on \ref{prob SDP SPCA} for two main reasons.
        From a theoretical perspective, stronger relaxations do not necessarily yield better \ref{prob SPCA} solutions in the input families studied in \cref{sec:rank_one,sec:adversarial pert}, the latter being a standard model in the statistics literature~\cite{AmiWai08,krauthgamer2015semidefinite,wang2016statistical}.
        From a computational perspective, \ref{prob SDP SPCA} can be (approximately) solved in seconds with our GPU implementation of the CGAL method~\cite{yurtsever2019conditional}, whereas most stronger relaxations might not be compatible with CGAL and thus scale poorly.
	\end{remark}

	Note that \cref{thm:spca rand alg} gives only a conditional expected approximation guarantee for \cref{alg:randomized spca simple}. This conditioning on the random event $\mathcal{R}$ does not substantially weaken the result, because $\mathcal{R}$ itself occurs with high probability. Hence, in practice, one expects $\mathcal{R}$ to hold in the vast majority of runs.
	    In practice, if one run of \cref{alg:randomized spca simple} has expected approximation ratio $\rho$, then Markov's inequality implies that a single run has ratio at most $2\rho$ with probability at least $1/2$. Consequently, the best of $N$ independent runs has ratio at most $2\rho$ with probability at least $1-2^{-N}$.
     
     \subsection{Assumptions on $W^*$ yielding small SSR}
     \label{sec:tech_cond}
     
     In this section, we discuss assumptions on $W^*$ that yield small $\textup{SSR}$, where $\textup{SSR}=\textup{SSR}(W^*)$ and $W^*$ is an (approximate) optimal solution to \ref{prob SDP SPCA}.
	 In the first proposition, we will show that if \ref{prob SDP SPCA} admits a sparse optimal solution $W^*$, then $\textup{SSR}(W^*)$ is upper bounded by the support of diagonal entries of $W^*$.
	 Then, we generalize this result to general low-rank optimal solutions, and show that $\textup{SSR}(W^*)$ is upper bounded by $\sqrt{rk}$, where $r:=\rk(W^*)$.
	 In this case, \cref{alg:randomized spca simple} gives an average approximation ratio of order $\mo(r\log{d})$.
	 We also note that this result could be very helpful in practice--it is possible to obtain a fixed-rank approximate solution to \ref{prob SDP SPCA} via CGAL~\cite{yurtsever2019conditional} with a fixed number of iterations and a low-rank initial primal solution, or to obtain a local low-rank solution to \ref{prob SDP SPCA} via the first-order approach proposed by Burer and Monteiro~\cite{burer2003nonlinear,burer2005local,cifuentes2021burer}, or via accelerated first-order methods~\cite{wang2025accelerated}.
	 Finally, we show that under certain assumptions, $\textup{SSR}(W^*)$ is small even when $W^*$ is of full rank. 
     We give our first proposition stating that a sparse $W^*$ has a small $\textup{SSR}(W^*)$:
	 \begin{proposition}
	 	\label{prop:sparse_solution_SSR}
	 	Let $W\succeq 0$, and assume that $\tr(W) = 1$.
	 	Define the diagonal support of $W$ as $\textup{DSupp}(W):=\{i\in [d]: W_{ii}\ne 0\}$.
	 	Then, $\textup{SSR}(W) \le \sqrt{\abs{\textup{DSupp}(W)}}$.
	 \end{proposition}
	 
	 \begin{prf}
	 	 	By Cauchy-Schwarz inequality, and by the fact that $\tr(W)=1$, $\textup{SSR}(W)$ is at most $\sqrt{|\textup{DSupp}(W)|}$.
	 \end{prf}
	 
	 Note that a solution $W^*$ with sparse diagonal support is by definition a low-rank matrix.
	 We now provide an upper bound for $\textup{SSR}(W)$ for general low-rank feasible solutions to \ref{prob SDP SPCA}:
	 \begin{proposition}
	 	\label{prop:low_rank_SSR}
	 	Let $W\succeq 0$, and assume that $\onorm{W}\le k$ and $\rk(W) = r$.
	 	Then, $\textup{SSR}(W) \le \sqrt{rk}$.
	 \end{proposition}
	 
	 \begin{prf}
	 	If $r = 0$, then $W = 0$ and the claim is trivial. 
		Hence, we may assume that $r\ge 1$.
	 	Suppose that $W = YY^\top$ with $Y\in\R^{d\times r}$. 
	 	Write $Y^\top = (y_1, y_2,\dots, y_d)$, with $y_i\in \R^r$.
	 	It is clear that $W_{ij} = y_i^\top y_j$ for any $i, j\in[d]$.
	 	For ease of notation, we define $r_i:= \norm{y_i}$. For every $i$ with $r_i > 0$, set $u_i:= y_i / r_i$; when $r_i = 0$, choose $u_i\in\R^r$ arbitrarily with $\norm{u_i}=1$.
	 	Then, we see that $\textup{SSR}(W) = \sum_{i = 1}^d r_i$ and $\onorm{W} = \sum_{i, j = 1}^d r_i r_j |u_i^\top u_j|$.

	 	We use a probabilistic viewpoint to lower bound $\onorm{W}$ by $\textup{SSR}(W)^2 / r$. 
	 	Let $I, J\in [d]$ be two i.i.d.~random variables with $\mp(I = i) = r_i / \textup{SSR}(W)$, and it is clear that
        {\setlength{\abovedisplayskip}{0pt}
  \setlength{\belowdisplayskip}{0pt}
	 	\begin{align*}
	 		\onorm{W} 
			= \textup{SSR}(W)^2 \me |u_I^\top u_J| 
			&\ge \textup{SSR}(W)^2 \me |u_I^\top u_J|^2 
            = \textup{SSR}(W)^2 \me (u_I^\top u_J) (u_J^\top u_I)\\
            & = \textup{SSR}(W)^2 \tr\left((\me u_I u_I^\top)^2\right),
	 	\end{align*}
        }
	 	where the inequality follows by the fact that $|u_I^\top u_J|\le 1$.        
	 	Define the matrix $G:=\me u_I u_I^\top\in\R^{r\times r}$, and let $\lambda_i(G)$ be the $i$-th largest eigenvalue of $G$, we obtain that
{\setlength{\abovedisplayskip}{2pt}
  \setlength{\belowdisplayskip}{0pt}
	 	\begin{align*}
	 		\tr(G^2) = \sum_{i = 1}^r \lambda_i(G)^2 = \frac{r \cdot \sum_{i = 1}^r \lambda_i(G)^2}{r} \ge \frac{\left(\sum_{i = 1}^r \lambda_i(G)\right)^2}{r}  = \frac{1}{r}.
	 	\end{align*}
        }
	 	via Cauchy-Schwarz inequality. Combining the last two displays, we get $\onorm{W}\ge \textup{SSR}(W)^2 / r$.
	 	Since $\onorm{W}\le k$, it follows that $\textup{SSR}(W)\le \sqrt{rk}$.
	 \end{prf}
	 
		 Finally, we show that, if the eigenvalues of $W^*$ decay exponentially, then $\textup{SSR}(W^*)^2=\mo(k\log{d})$:
		 \begin{proposition}
	 		\label{prop:exponential_decay_SSR}
	 		Let $W\succeq 0$ and assume that $\tr(W)=1$ and $\onorm{W}\le k$. 
			Suppose that there exists a constant $q \in (0,1)$ such that the $i$-th largest eigenvalue of $W$, i.e., $\lambda_{i}(W)$, satisfies that $\lambda_{i}(W) \le q^{i-1} \cdot \lambda_{1}(W)$ for all $i\in [d]$.
	 		Then, there exists an absolute constant $C>0$ such that $\textup{SSR}(W) \le C\sqrt{k\log{d} / \log\left(1 / q\right)}$.
		 \end{proposition}
	 
	 \begin{prf}
	 	Let $W\succeq 0$ and assume that $\tr(W) = 1$ and $\onorm{W}\le k$.
	 	Let the singular value decomposition of $W$ be $\sum_{i = 1}^d \lambda_i(W) u_i u_i^\top$. 
	 	Define the orthogonal matrix $U:= (u_1, u_2, \dots, u_d)$, and thus $U^{-1} = U^\top$ and hence $UU^\top = \sum_{i = 1}^d u_i u_i^\top = I_d$.
	 		In other words, we have that $\sum_{i = 1}^d (u_i)_j^2 = 1$ for any $j \in [d]$.
	 	Let $r$ be the smallest integer such that $q^r \le 1 / d^2$, and thus $r = \lceil 2\log{d} / \log(1 / q) \rceil$.
	 	Write $W_1:= \sum_{i = 1}^r \lambda_i(W) u_i u_i^\top$, and $W_2:= W - W_1 = \sum_{i = r+1}^d \lambda_i(W) u_i u_i^\top$.
	 	It is clear that 
        {\setlength{\abovedisplayskip}{2pt}
  \setlength{\belowdisplayskip}{0pt}
	 	\begin{align*}
	 		\textup{SSR}(W) = \sum_{j=1}^d \sqrt{(W_1)_{jj} + (W_2)_{jj}} \le \sum_{j = 1}^d \left(\sqrt{(W_1)_{jj}} + \sqrt{(W_2)_{jj}}\right).
	 	\end{align*}
        }
	 	For any $j,\ell\in[d]$, we have
        {\setlength{\abovedisplayskip}{2pt}
  \setlength{\belowdisplayskip}{0pt}
	 	\begin{align*}
	 		\abs{(W_2)_{j\ell}}
	 		&= \abs{\sum_{i = r+1}^d \lambda_i(W) (u_i)_j (u_i)_\ell}
	 		\le \lambda_{r+1}(W) \sum_{i = r+1}^d \abs{(u_i)_j (u_i)_\ell} \\
	 		&\le q^r \Big(\sum_{i = r+1}^d (u_i)_j^2\Big)^{1/2} \Big(\sum_{i = r+1}^d (u_i)_\ell^2\Big)^{1/2}
	 		\le \frac{1}{d^2},
	 	\end{align*}
        }
	 	where we used $\lambda_{r+1}(W)\le q^r\lambda_1(W)\le q^r$ and $\lambda_1(W)\le \tr(W)=1$.
	 	Therefore, $\onorm{W_2}\le d^2\cdot d^{-2}=1$, and hence
        {\setlength{\abovedisplayskip}{2pt}
  \setlength{\belowdisplayskip}{0pt}
	 	\begin{align*}
	 		\onorm{W_1} \le \onorm{W} + \onorm{W_2} \le k + 1 \le 2k,
	 	\end{align*}
        }
	 	since $k$ is a positive integer. As $\rk(W_1)\le r$, \cref{prop:low_rank_SSR} yields
        {\setlength{\abovedisplayskip}{2pt}
  \setlength{\belowdisplayskip}{0pt}
	 	\begin{align*}
	 		\sum_{j=1}^d \sqrt{(W_1)_{jj}} \le \sqrt{2rk}.
	 	\end{align*}
        }
	 	For the term involving $W_2$, we see that $\sqrt{(W_2)_{jj}} = \sqrt{\sum_{i = r+1}^d \lambda_{i}(W) (u_i)_j^2} \le \sqrt{1 /d^2} = 1 / d$.
	 	Therefore, we obtain that $\sum_{j=1}^d \sqrt{(W_2)_{jj}} \le d\cdot \frac{1}{d} = 1$, and thus
        {\setlength{\abovedisplayskip}{2pt}
  \setlength{\belowdisplayskip}{0pt}
	 	\begin{align*}
	 		\textup{SSR}(W) \le \sqrt{2rk} + 1.
	 	\end{align*}
        }
	 	Since $r = \lceil 2\log{d} / \log(1 / q) \rceil$, we are done.
	 \end{prf}

     \subsection{Sufficient conditions for a rank-one optimal solution}
     \label{sec:rank_one}
     In this section, we provide further understanding for the technical condition~\eqref{eqn:spca tech cond}, by providing sufficient conditions for obtaining a rank-one optimal solution to \ref{prob SDP SPCA}.
     We note that while conditions of this type have been investigated in the context of general QCQPs~\cite{wang2022tightness}, in this section we focus on conditions specifically tailored to \ref{prob SDP SPCA}.
     By \cref{prop:low_rank_SSR}, $\textup{SSR}(W^*)$ is upper bounded by $\sqrt{k}$ when there is a rank-one (approximate) optimal solution to \ref{prob SDP SPCA}. 
     To our knowledge, we give the first deterministic classes of instances for \ref{prob SDP SPCA} that guarantee rank-one optimal solutions.
     
     We start by stating a simple fact that, if the input matrix $A$ admits a maximum eigenvector with an $\ell_1$-norm upper bounded by $\sqrt{k}$, then \ref{prob SDP SPCA} admits a rank-one optimal solution.
     
     \begin{fact}
     	\label{fact:sparse_max_eigenvec}
     	Let that $A\succeq 0$, and denote by $v_1(A)$ an eigenvector corresponding to the maximum eigenvalue of $A$ with $\norm{v_1(A)} = 1$.
     	Assume that $\onorm{v_1(A)} \le \sqrt{k}$, then $v_1(A)v_1(A)^\top$ is an optimal solution to \ref{prob SDP SPCA}.
     \end{fact}
     
     \begin{prf}
     	Let $W\succeq 0$ be a feasible solution to \ref{prob SDP SPCA}, and assume that its singular value decomposition is $W = \sum_{i = 1}^d \lambda_i v_i v_i^\top$. 
     	Denote by $\lambda_{1}(A)$ the largest eigenvalue of $A$.
     	It is clear that 
        {\setlength{\abovedisplayskip}{2pt}
  \setlength{\belowdisplayskip}{0pt}
     	\begin{align*}
     		\tr(AW) = \sum_{i = 1}^d \lambda_i v_i^\top A v_i \le \sum_{i = 1}^d \lambda_i \cdot \lambda_1(A) = \lambda_{1}(A),
     	\end{align*}
        }
     	where the last equality follows from the fact that $\tr(W) = 1$.
     	We obtain our desired result by noticing the fact that $\onorm{v_1(A)v_1(A)^\top} = \onorm{v_1(A)}^2 \le k$ and $\tr(A v_1(A)v_1(A)^\top) = \lambda_1(A)$.
     \end{prf}
     
     In the next example, we provide a class of instances that satisfy the assumptions in \cref{fact:sparse_max_eigenvec}:
     
     \begin{example}
     	\label{example:sparse_eigenvec}
     	Suppose that the matrix $A = \lambda I - S \succeq 0$ for some $\lambda \ge 0$ and $S\succeq 0$.
     	Denote by $\{n_i\}_{i = 1}^r$ an orthonormal basis of the nullspace of $S$ with dimension $r\le d$, and denote by $N:=(n_1, n_2, \dots, n_r)$.
     	For $i\in [d]$, let $N_i\in\R^r$ be the row vector of $N$.
     	Assume that $\sum_{i = 1}^d \norm{N_i} \le \sqrt{rk}$, then $A$ admits a top eigenvector $v$ such that $\onorm{v}\le \sqrt{k}$.
     \end{example}
     
     \begin{prf}
     	It only suffices to show that there exists a unit vector $x\in \R^r$ such that $\sum_{i = 1}^d \abs{\sum_{j = 1}^r N_{ij} x_j} \le \sqrt{k}$.
     	Indeed, if such $x$ exists, the vector $Nx$ is by definition a top eigenvector of $A$ with $\norm{Nx} = 1$ and $\onorm{Nx} \le \sqrt{k}$.
     	
     	We use a probabilistic method to prove the desired result. 
     	Let $\{\epsilon_i\}_{i = 1}^r$ be i.i.d.~Rademacher random variables, i.e., $\mp(\epsilon_i = \pm 1) = 1/2$.
     	By Khintchine inequality~\cite{Haagerup1981}, we obtain that $\me \left[\abs{\sum_{j = 1}^r N_{ij} \epsilon_j}\right] \le \norm{N_i}$, and therefore $\me \sum_{i = 1}^d \abs{\sum_{j = 1}^r N_{ij} \epsilon_j} \le \sum_{i = 1}^d \norm{N_i} \le \sqrt{rk}$.
     	Hence, we see that there exists a vector $y \in \{\pm 1\}^r$ such that $\sum_{i = 1}^d \abs{\sum_{j = 1}^r N_{ij} y_j} \le \sqrt{rk}$. Taking $x = y / \sqrt{r}$ concludes the proof. 
     \end{prf}
     
     However, in practice, the assumptions in \cref{fact:sparse_max_eigenvec} might not hold.
     In the remainder of the section, we provide other classes of instances that allow the top eigenvector of $A$ to have a larger $\ell_1$-norm, yet still guarantees that \ref{prob SDP SPCA} admits a rank-one optimal solution.
     In the next theorem, we show that if $A$ is the sum of a non-negative multiple of the identity and a rank-one matrix, then \ref{prob SDP SPCA} admits a rank-one solution under mild assumptions:
     \begin{theorem}
     	\label{thm:rank_one_input}
     	Assume that $A = \lambda I_d + uu^\top$ for some vector $u\in \R^d$ with $m:=\onorm{u} / \norm{u} > \sqrt{k}$.
     	Let $T:=\supp(u)$, and assume that
        {\setlength{\abovedisplayskip}{2pt}
  \setlength{\belowdisplayskip}{0pt}
     	\begin{align}
        \label{eqn:ass rank one}
     	\frac{m - \sqrt{k\cdot \frac{|T| - m^2}{|T| - k}}}{|T|} < \min_{i\in T} \abs{u_i} < \frac{m + \sqrt{k\cdot \frac{|T| - m^2}{|T| - k}}}{|T|}.
     	\end{align}
        }
     	Then, \ref{prob SDP SPCA} admits a unique optimal solution $w^* (w^*)^\top$ with $\supp(w^*) = T$.
     \end{theorem}
    
     We note that if $A$ is of a fixed rank $r$ (up to an addition of a scaling of identity), an optimal solution to \ref{prob SPCA} could be found in polynomial time~\cite{papailiopoulos2013sparse,del2022sparse}.
     In the next proposition, we provide a special case in which we can obtain a rank-one optimal solution without matrix $A$ having a fixed rank. 
	 We then build on this case to develop a more general result in \cref{thm:rank-one}.

\begin{restatable}{proposition}{proprankoneexactsign}
\label{prop:rank_one_exact_sign}
Assume $A\succeq 0$ and let $S\subseteq[d]$ satisfy $|S|=k$.
Let $B:=A_{S,S}\in\R^{k\times k}$ have eigenvalues $\lambda_1>\lambda_2\ge\cdots\ge\lambda_k\ge 0$, and let $v_1$ be a unit eigenvector associated with $\lambda_1$.
Assume that $v_1 = s / \sqrt{k}$ for some $s \in \{\pm1\}^k$ and that
\begin{equation}
\label{eqn:off-support_upper_bound}
  \max\{\max_{i\in S, j\in S^c}|A_{ij}|,\ \max_{i,j\in S^c}|A_{ij}|\}
  \le \frac{\lambda_1-\lambda_2}{2k}.
\end{equation}
Then \ref{prob SDP SPCA} admits a rank-one optimal solution $W^* = w^*(w^*)^\top$, where $w^*\in\R^d$ is given by
\begin{equation*}
  (w^*)_S=\frac{s}{\sqrt{k}},
  \qquad
  (w^*)_{S^c}=0.
\end{equation*}
Moreover, if the inequality \eqref{eqn:off-support_upper_bound} is strict, then $W^*$ is the unique optimal solution.
\end{restatable}

We now turn to the general case, where the leading eigenvector need not
be a scaling of a $\{\pm 1\}$-vector.
The assumption can be packaged as follows:

\begin{assumption}
\label{ass:rank-one}
Assume $A\succeq 0$ and let $S\subseteq[d]$ be nonempty. 
Write $B:=A_{S,S}\in\R^{|S|\times |S|}$ and let $\lambda_1>\lambda_2\ge\cdots\ge\lambda_{|S|}\ge 0$ be its eigenvalues.
Let $v_1$ be a unit eigenvector for $\lambda_1$, assume $v_1$ has no zero coordinates, and set $s:=\sign(v_1)$, $m:=\min_i |(v_1)_i|>0$, and $\alpha:=\onorm{v_1}\,m$. 
Let $P_2$ be the orthogonal projection onto the $\lambda_2$-eigenspace, and for each eigenvalue $\theta<\lambda_2$ let $P_\theta$ be the orthogonal projection onto the $\theta$-eigenspace. 
Define
\begin{equation*}
  \begin{aligned}
    M &:= \left(\sum_{\theta<\lambda_2}
      \frac{\norm{P_\theta s}^2}{(\lambda_2-\theta)^2}\right)^{1/2},
    &\qquad
    w_2 &:= \frac{\onorm{v_1}}{\lambda_1-\lambda_2}v_1
      + \sum_{\theta<\lambda_2}\frac{P_\theta s}{\theta-\lambda_2}, \\
    R(\lambda) &:= \frac{\|(B-\lambda I)^{-1}s\|_1}{\|(B-\lambda I)^{-1}s\|_2},
    &\qquad
    \eta &:= \max\{\max_{i\in S, j\in S^c}|A_{ij}|,\ \max_{i,j\in S^c}|A_{ij}|\},
  \end{aligned}
\end{equation*}
for $\lambda\in(\lambda_2,\lambda_1)$.
Assume:
\begin{enumerate}[label=A\arabic*, ref=A\arabic*]

  \item \label{item:ass_resolvent_sign}
  (Orthogonality and lower-spectrum control)
  $P_2 s=0$ and $M<\alpha/(\lambda_1-\lambda_2)$.

  \item \label{item:ass_resolvent_endpoints}
  (Endpoint comparison for $R$)
  $\big(\|w_2\|_1/\|w_2\|_2-\sqrt{k}\big)
  \big(\|v_1\|_1/\|v_1\|_2-\sqrt{k}\big)<0$.

  \item \label{item:ass_resolvent_window}
  (Interior level and spectral margin)
  There exists $\lambda^*\in(\lambda_2,\lambda_1)$ such that
  $R(\lambda^*)=\sqrt{k}$ and
  $(\lambda^* - \lambda_2)\big(\alpha/(\lambda_1-\lambda^*)-M\big)\ge 1$.

  \item \label{item:ass_resolvent_offsupport}
  (Off-support bound at $\lambda^*$)
  With $\mu^* := 1/(\sqrt{k}\,\|(B-\lambda^* I)^{-1}s\|_2)$, one has
  $\eta\le \mu^*$.
\end{enumerate}
\end{assumption}

\begin{remark}[How to understand Assumption~\ref{ass:rank-one}]
\label{rmk:resolvent_reading}
This assumption captures a setting in which the support block $B=A_{S,S}$ already exhibits a dominant sign pattern on $S$, the associated resolvent ratio reaches the target level $\sqrt{k}$ at a shift with sufficient spectral margin, and the interaction with $S^c$ stays uniformly small. 
With $R$ as above, \ref{item:ass_resolvent_sign} controls the lower-spectrum contribution to $s=\sign(v_1)$, \ref{item:ass_resolvent_endpoints} forces $R$ to cross $\sqrt{k}$, \ref{item:ass_resolvent_window} chooses a crossing point with enough margin, and by \cref{lem:q4_quadratic} the quadratic inequality in that item automatically holds for any such $\lambda^*$ sufficiently close to $\lambda_1$. 
Finally, \ref{item:ass_resolvent_offsupport} bounds the entries outside $S$. 
Consequently, once a candidate support set $S$ is given, verifying Assumption~\ref{ass:rank-one} reduces to an eigendecomposition of $B=A_{S,S}$, a one-dimensional root-finding problem for $R(\lambda)=\sqrt{k}$ on $(\lambda_2,\lambda_1)$, and a direct entrywise comparison outside $S$.

We note that this does not remove the difficulty of finding $S$ in full generality. 
Its value is that whenever structure or preprocessing shrinks the candidate family, possibly still exponentially but far below the full search over subsets, each candidate can be certified explicitly. 
Examples include threshold-separated blocks, where a threshold can isolate $S$ as a connected component in the spirit of covariance thresholding \cite{deshpande2014sparse}; block-diagonal or nearly block-diagonal matrices, where one searches only over blocks or short lists of merged blocks as in \cite{delpia2025efficient}; and ordered or grouped variables, where contiguous supports $S=\{a,\ldots,b\}$ yield only $d(d+1)/2$ possibilities (or $d-k+1$ when $|S|=k$ is fixed), while unions of at most $r$ prescribed groups among $G$ groups yield $O(G^r)$ candidates for fixed $r$. 
In such regimes, the assumption serves as an explicit certificate for supports found either by a polynomial-time screening rule or by a reduced exact search.
\end{remark}

We are now ready to state the general resolvent-based result:

\begin{restatable}{theorem}{thmrankone}
\label{thm:rank-one}
Under Assumption~\ref{ass:rank-one}, \ref{prob SDP SPCA} admits a rank-one optimal solution $W^* = w^*(w^*)^\top$, with $w^*\in \R^d$ and $\supp(w^*) = S$.
Moreover, if the spectral-margin inequality in \ref{item:ass_resolvent_window} is strict, then $W^*$ is the unique optimal solution to \ref{prob SDP SPCA}.
\end{restatable}

For a rank-one optimal solution $W^*$, we can apply \cref{alg:multi_ra} for a set number of iterations to obtain a solution to \ref{prob SPCA} with an approximation ratio of $\mo(\log{d})$, as guaranteed by \cref{thm:spca rand alg}.
However, we acknowledge that, when $W^*$ is rank-one, stronger approximation guarantees are achievable, as shown in \cite{dey2017sparse,dey2022solving}. 
The authors demonstrate that solving specific non-convex quadratic programs, equivalent to enforcing a rank-one solution in \ref{prob SDP SPCA}, yields a solution with a constant approximation ratio, using a slightly different sampling rule compared to \cref{alg:randomized spca simple}.
Despite some algorithmic overlap, our approach diverges significantly from theirs in several ways: (i) \cref{alg:randomized spca simple} takes a square positive semidefinite input, while \cite{dey2017sparse,dey2022solving} operates on vector solution(s); (ii) this distinction also leads to different proof techniques: \cite{dey2017sparse,dey2022solving} analyze the size of feasible region, whereas we derive good-quality solutions based on properties of (sub-)Gaussian variables, as will be made clear in \cref{sec:proof of ra}; and (iii) while their work focuses on efficiently solving non-convex programs (except exponential time complexity in the worst case), we aim to develop a polynomial-time algorithm with strong approximation ratios and identify input classes where even better performance can be achieved, as discussed in this and the following section.

\section{Robustness of basic SDP relaxation within a general covariance model}
	\label{sec:adversarial pert}
 
	In this section, we study a statistical model where the input matrix $A$ is of the form $A = (B + M)^\top (B + M)$, where $B \in \R^{n \times d}$ is a data matrix with a certain sparse signal, and $M$ is a bounded adversarial perturbation matrix. 
    Here, $n$ is known to be the number of samples.
    One goal of this section is to demonstrate that under this model assumption, $W^*$ closely approximates a sparse signal embedded within the model. 
    Consequently, $W^*$ can be regarded as an effective approximation of the true sparse signal.
    After that, we show that in this model, \cref{alg:multi_ra} achieves an approximation ratio close to one, with the near-one guarantee coming from the deterministic solution built from the largest diagonal entries of $W^*$.

	In \cite{d2020sparse}, the authors studied the spiked Wishart model, where $M$ is a zero matrix, and $B$ is a spiked standard Gaussian matrix, i.e., every row of $B$ is an i.i.d.~random vector drawn from $\mathcal{N}(0_d, I_d + \beta v v^\top)$, with $\beta > 0$ and $v$ being a $k$-sparse vector. 
	The authors demonstrated that an optimal solution to \ref{prob SDP SPCA} provides a good approximation to $v$ for a certain sample size $n$, which in that model is roughly $\mo(k\log{d})$ since $M=0$.
	However, such spiked assumption is often not applicable to real-life scenarios, as it is uncommon for each row of the actual data matrix $B$ to represent a sum of a sparse signal realization $\sqrt{\beta} v$ and independent standard Gaussian noise. 
    This discrepancy serves as the impetus for exploring the performance of \ref{prob SDP SPCA} in more generalized contexts. 
    These contexts are characterized not only by the presence of sub-Gaussian random variables but also by the inclusion of multiple realizations of signals, amongst which a sparse dominant signal is present. 
    Formally, we introduce \cref{model:adversarial}:
	\begin{model}
		\label{model:adversarial}
		The input matrix $A$ can be written as $A = (B + M)^\top (B+M)$, where $B\in\R^{n\times d}$ is a random matrix with i.i.d.~sub-Gaussian rows with parameter $\sigma^2$, and $M$ is a modification matrix such that its maximal column norm is upper bounded by a constant $b>0$, i.e., $\|M\|_{1\rightarrow2}\le b$. 
		Furthermore, the rows of $B$ have zero means and admit a covariance matrix $\Sigma$, such that $\Sigma$ has a $k$-sparse maximal eigenvector $v$ associated with eigenvalue $\lambda_1$.
	\end{model}

    We present in \cref{thm:randomized_alg_in_stat_model} that, given a sufficient number of samples $n$, the deterministic solution in our algorithm already yields an approximation ratio close to one.
    This result develops on the following characterization that $W^*$ is close enough to $vv^\top$, implying that \ref{prob SDP SPCA} is robust to adversarial perturbations in \cref{model:adversarial}:

    \begin{restatable}{proposition}{proprobustrandom}
		\label{prop:robust random}
		In \cref{model:adversarial}, denote $\lambda_1, \lambda_2$ to be the largest and second largest eigenvalue of $\Sigma$, respectively, and assume $\lambda_1 - \lambda_2 > 0$.
        Let $v$ be the eigenvector associated with $\lambda_1$, and denote $a:=\min_{i: v_i \ne 0} |v_i|$.
        Let $W^*$ be an optimal solution to \ref{prob SDP SPCA}. 
        Then, there exists an absolute constant $C^* > 0$ such that when $n$ is greater or equal to
        {\setlength{\abovedisplayskip}{0pt}
  \setlength{\belowdisplayskip}{0pt}
            \begin{align}
            \label{eqn:def_of_n_star}
                n^*:= \max\Big\{C^* \cdot \Big[\frac{k^2 \sigma^4\log{d} + b^2 k^2 \Big(\sigma^2 + \max \Sigma_{ii} \Big)}{(\lambda_1 - \lambda_2)^2 a^4} + \frac{kb^2}{(\lambda_1 - \lambda_2) a^2} \Big], \frac{4}{a^2}, \log{d} \Big\},
            \end{align}
            }
            then $\infnorm{W^* - vv^\top} \le {a^2}/{2}$ holds with probability at least $1 - d^{-10}$.
	\end{restatable}

 \begin{remark}
		In this remark, we discuss the sample complexity required for \ref{prob SDP SPCA} to recover $\supp(v)$ in \cref{model:adversarial}, which might be of independent interests. 
		According to \cref{prop:robust random}, given a fixed signal intensity $\lambda_1 - \lambda_2$, a fixed variance factor $\sigma^2$, and a fixed $a$, if 
        {\setlength{\abovedisplayskip}{0pt}
  \setlength{\belowdisplayskip}{0pt}
		\begin{align}
			\label{eqn:sample complexity}
			 n = \Omega\left( k^2\log{d} + k^2 b^2 \lambda_1 + kb^2\log{d}\right),
		\end{align}
        }
	        one can recover $\supp(v)$ with high probability via \ref{prob SDP SPCA}.
		We note that the term $\Omega(k^2\log{d})$ in \eqref{eqn:sample complexity} is due to the recovery of $\supp(v)$ in \cref{model:adversarial} without any adversarial perturbation, which is consistent with the findings of \cite{wang2016statistical}. 
		The term $\Omega(k^2 b^2 \lambda_1 + kb^2\log{d})$ in \eqref{eqn:sample complexity} reflects the additional number of samples required to recover $\supp(v)$ under adversarial perturbations.
	        Finally, this sample-complexity bound alone does not imply \eqref{eqn:spca tech cond}. Indeed, \cref{prop:robust random} controls $\|W^*-vv^\top\|_\infty$, but that control by itself is not strong enough to deduce the required logarithmic bound on $\textup{SSR}$.
		\end{remark}

    We are ready to develop the main theorem in this section. 
    In contrast to the approximation guarantees from \cref{sec:app_guarantee}, which are driven by the randomized solutions in \cref{alg:multi_ra}, the core message here is that the deterministic solution from lines 1--2 of \cref{alg:multi_ra} already finds a solution with approximation ratio near one.
    \begin{restatable}{theorem}{thmrainstatmodel}
    \label{thm:randomized_alg_in_stat_model}
    In \cref{model:adversarial}, denote $\lambda_1, \lambda_2$ to be the largest and second largest eigenvalue of $\Sigma$, respectively, and assume $\lambda_1 - \lambda_2 > 0$. 
    Let $W^*$ be an optimal solution to \ref{prob SDP SPCA}. 
    Denote $n^*$ the number defined in \eqref{eqn:def_of_n_star}.
    Then, for $l\ge 1$, suppose that one has $n\ge l\cdot n^*$, then for any number of iterations $N \ge 0$, \cref{alg:multi_ra} with input $(A, W^*, k, N)$ has an approximation ratio of at most $1 + 2 / (8\sqrt{l} - 1)$ with probability at least $1 - d^{-10}$.
    \end{restatable}
\section{Numerical tests}
	\label{sec:numerical tests spca}

	In this section, we present our numerical results.
	Our main objective is to evaluate the performance of \cref{alg:multi_ra} on real-world datasets and compare computational performance with existing state-of-the-art algorithms  that run in \emph{polynomial time} and have an approximation guarantee.
    It should be noted that, while comparisons are made, they are limited to a selected subset of existing algorithms, rather than an exhaustive review of all available methods.
    For further comparisons of \ref{prob SDP SPCA} with other algorithms, readers are directed to the recent study in \cite{chowdhury2020approximation}.
	We conducted tests on several real-world datasets, with the dimension of matrix $A$ ranging from 79 to 2000. 
	For most available SDP solvers, including Mosek~\cite{mosek} and SCS~\cite{scs}, face weak scalability, making it hard for them to handle semidefinite programs with dimensions exceeding 1000. 
	As a result, we use CGAL~\cite{yurtsever2019conditional} to obtain approximation solutions to \ref{prob SDP SPCA}.
    It is worth noting that, in \cref{alg:randomized spca simple}, we simply take the set $T$ (on line 8) with $|T| = k - |S|$ to be the index set in $[d]\backslash S$ that yields largest $W_{ii}$'s (with ties broken arbitrarily).
    
    \paragraph{Introduction to CGAL.}
    The conditional gradient augmented Lagrangian framework
(CGAL)~\cite{yurtsever2019conditional} is an iterative algorithm that (approximately) solves the following problem: 
    {\setlength{\abovedisplayskip}{0pt}
  \setlength{\belowdisplayskip}{0pt}%
	\begin{align*}
		f(x^*) := \min  \ f(x) \quad
		\text{s.t. }  x\in \mathcal{X}, \ C x \in \mathcal{K},
	\end{align*}
    }
	where $f$ is a convex and $L$-smooth function, $C$ is a linear mapping, $\mathcal{X}$ is a convex compact set and $\mathcal{K}$ is a convex set. 
	In the $m$-th iteration, CGAL is guaranteed to find $x_m$ such that $|f(x_m) - f(x^*)| \le \mo(m^{-1/2})$ and $\text{dist}(C x_m, \mathcal{K}) \le \mo(m^{-1/2})$.
	In the tests, number of iterations in CGAL is set to $100$ and the parameter $\lambda_0$ set to $1$, and we initialize $x_0$ to be the zero matrix---which guarantees that $\rk(x_m)\le m$.
    We find that this algorithm could be efficiently implemented on GPU.

    \paragraph{Hardware.}
    We conducted all tests on a personal computer with 8 Cores i7-9700K 3.60GHz CPU, 64 GB of memory, and NVIDIA GEFORCE RTX 2080 SUPER with 8 GB of GPU memory.

    \paragraph{Baselines.}
    We compare polynomial-time algorithms studied in \cite{papailiopoulos2013sparse,chan2016approximability,li2020exact,berk2019certifiably}. 
    It is worth noting that Chan's algorithm studied in \cite{chan2016approximability} finds a $\min\{\sqrt{k},d^{1/3}\}$-approximate solution to \ref{prob SPCA}.
    To the best of our knowledge, this is the best known approximation ratio that can be obtained in polynomial time for general \ref{prob SPCA}.
    We also compare the performance of \cref{alg:multi_ra}, among the Greedy algorithm and the Local Search algorithm studied in \cite{li2020exact}, and the Low-Rank method studied in \cite{papailiopoulos2013sparse}.
    The Greedy algorithm and the Local Search algorithm both find $k$-approximate solutions in polynomial time, while the Low-Rank method~\cite{papailiopoulos2013sparse} finds a solution with approximation ratio depending on the decay of the eigenvalues of the input matrix $A$.
    For computational efficiency, we apply the Low-Rank method with the rank-2 approximation to $A$.
    
    We also compare with the Branch-and-Bound (BB) algorithm  from \cite{berk2019certifiably} and MSPCA algorithm from \cite{dey2022solving}, while the latter finds \ref{prob SPCA} both a lower bound (LB) using a heuristic and an upper bound (UB) using MILP.
    Although BB and MSPCA frequently find good solutions within our time limit, both algorithms have worst-case running times that could grow exponentially with $k$, so we treat their results as supplementary to the core comparison among polynomial-time methods. 
    BB has no approximation guarantee, and MSPCA, while returning an upper bound within a constant factor of the optimum, still provides no polynomial runtime bound. 
    By contrast, Greedy, Local Search, Low-Rank, and \cref{alg:multi_ra} combine provable approximation ratios with guaranteed polynomial complexity, making them the efficient choice for large-scale instances (see \cref{tab:full_results}).

    \paragraph{Summary of results.}
    Across 41 benchmark instances, and up to a tolerance of $10^{-3}$ in objective values, \cref{alg:multi_ra} (RA) attains the best objective in 31 cases and matches or exceeds the competitors as follows.
    Runtime comparisons are based on average runtime (\cref{tab:full_results}).
    \begin{itemize}[leftmargin=2em]
      \item Greedy algorithm: 85\% of instances (20\% strictly better); ~2.2x faster than RA.
      \item Local Search algorithm: 80\% (10\% strictly better); ~2.6x slower than RA.
      \item Low-Rank method: 90\% (39\% strictly better); ~27.4x slower than RA.
      \item Chan’s algorithm: 95\% (78\% strictly better); ~4.3x faster than RA.
      \item BB: 75\% (2\% strictly better); ~129.5x slower than RA.
      \item MSPCA LB: 90\% (44\% strictly better); ~41.0x slower than RA.
    \end{itemize}

    In \cref{fig:gap-grid}, we report the following \emph{Chan-normalized Gap} for the five algorithms, and use Chan's Algorithm as a baseline:
    \begin{equation*}
        \textup{Chan-normalized Gap} := \frac{\textup{Obj}_{\textup{alg}} - \textup{Obj}_{\textup{Chan}}}{\textup{Obj}_{\textup{Chan}}} \times 100\%,
    \end{equation*}
    where $\textup{Obj}_{\textup{Chan}}$ is the \ref{prob SPCA} objective value found by Chan's algorithm, and $\textup{Obj}_{\textup{alg}}$ is that of the algorithm under test. 
    Relative to Chan’s method, RA shows an average gap of 0.34\% (max 2.80\%, min -1.42\%).
    The average Chan-normalized gaps for the Greedy algorithm, the Local Search algorithm, the Low-Rank method, BB, and MSPCA LB are -0.16\%, -0.03\%, -4.16\%, 0.38\%, and -2.43\%, respectively. 
    Note that RA obtains the highest average Chan-normalized gap among all polynomial-time algorithms.
    In six instances (14\% of all instances) RA gains over 1\% compared to Chan's Algorithm.
    We also notice that, although the gaps for Greedy, Local Search, and RA shown in \cref{fig:gap-grid} are oftentimes close to each other, RA is able to find better solutions when they fail. 
    For example, for the instance $k=5$ on the LymphomaCov dataset, both Greedy algorithm and Local Search algorithm find solutions with gaps around -11.97\%, yet RA finds a solution with a gap 0.07\%.
    Although BB often attains equal or better objective values compared to all four polynomial-time methods, its Chan-normalized gap exceeds that of RA by only 0.04\% on average and at most 1.43\%.
    
    \begin{figure}[t]
    		\centering
    		\includegraphics[width=\textwidth]{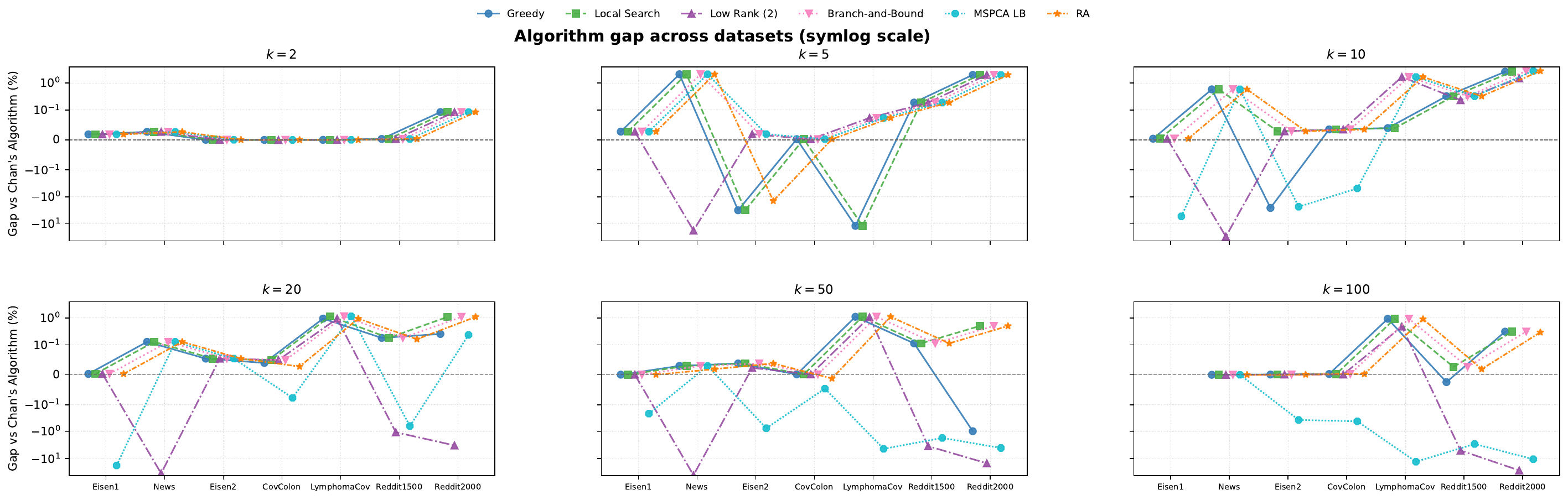}
    		\caption{\textbf{Chan-normalized gaps (higher = better).} Each panel fixes the sparsity level~$k$ and plots the Chan-normalized gaps of the six algorithms across the seven benchmark datasets, ordered from the smallest to the largest dimension (left to right).
    		Our Randomized Algorithm achieves the best objective on 31 of the 41 instances (75\%) and never falls below Chan by more than 1.5\%. Note that we do not report the result for $k = 100$ on the Eisen1 dataset as its dimension is smaller than 100.}
    		\label{fig:gap-grid}
   	\end{figure}
   	
   	In \cref{fig:runtime-grid}, we report the runtimes of different algorithms.
   	It highlights that RA maintains these accuracy gains at practical speeds: the median runtime is merely 2.57 seconds, and maximum runtime is 11.95 seconds.
   	It should be noted that the sampling of $N = 3000$ solutions only takes less than 3.5 seconds for 75\% of the instances and the maximum sampling time among the 41 instances is 9.06 seconds.
    RA is approximately 312 times faster than BB on average, 1512 times faster in the best case ($k=20$ on Eisen2 dataset). 
   	Moreover, RA is approximately 15 times faster than Low-Rank method on average, 67 times faster in the best case ($k=100$ on LymphomaCov dataset). 
    MSPCA’s runtime covers both its lower and upper bound computations and is included only for completeness.
    On all 20 instances with $k\ge 20$, RA is 2.5 times faster than the Local Search algorithm on average, and 11 times faster in the best case ($k=100$ on Reddit1500 dataset). 
   	Although RA is in general slower than the Greedy algorithm and Chan's algorithm, it offers an accuracy–efficiency trade-off, coupling better objectives with practical scalability for large SPCA instances, as shown in \cref{fig:gap-grid}.
	
	\begin{figure}[htp]
		\centering
		\includegraphics[width=\textwidth]{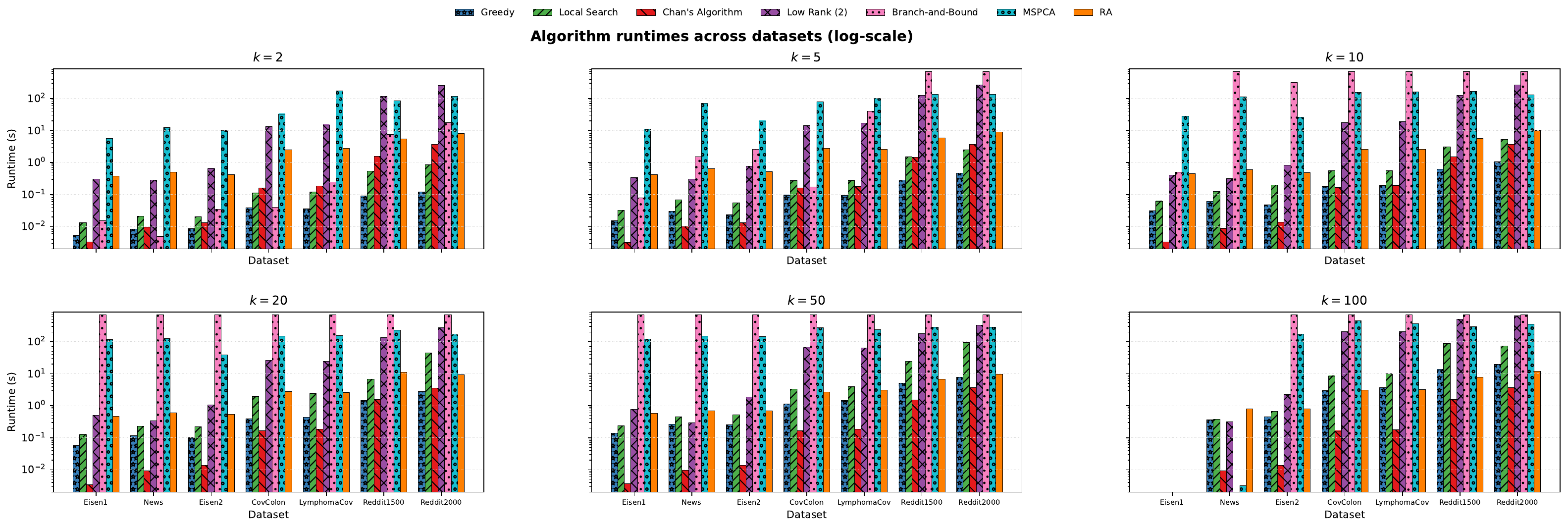}
		\caption{\textbf{Wall-clock runtimes (log scale).}
			The same experiments as \cref{fig:gap-grid} but reporting runtime.  
			RA terminates in under 10 seconds on 90\% of instances.  
			It is roughly 312 times faster than BB and 15 times faster than the Low-Rank method, while only one order of magnitude slower than the Greedy and Chan's algorithm.
			When $k\ge 20$, the runtime of RA is on average 2.5 times faster than the Local Search algorithm.
			We do not report the result for $k = 100$ on the Eisen1 dataset as its dimension is less than 100.}
		\label{fig:runtime-grid}
	\end{figure}

	    It is important to note that, across all instances the assumption \eqref{eqn:spca tech cond} holds with $c_0 \le 2.16$ on average, the 80-th percentile for $c_0$ is merely 2.21, while the 90-th percentile is 5.29.
	    The two clearest outliers are the CovColon and LymphomaCov instances with $(d,k)=(500,2)$, where $c_0$ equals 13.003 and 10.030, respectively. Since $c_0 = \textup{SSR}/\sqrt{k}$, these values indicate that the diagonal of the SDP solution is unusually diffuse at the smallest sparsity level. On the same two datasets, however, $c_0$ drops quickly as $k$ increases (to 7.487 and 4.933 already at $k=5$, and below 3.6 by $k=20$), suggesting that this spike is specific to the extreme $k=2$ regime rather than a broader instability of the relaxation. Importantly, \cref{alg:multi_ra} still attains the best objective value on both instances.
	    As a result, on these benchmark instances and under the verified condition \eqref{eqn:spca tech cond}, RA has an $\mo(\log{d})$-approximation guarantee.
	    To connect this observation to the theory, we also inspect the empirical ratio between the objective value returned by \cref{alg:multi_ra} and the SDP objective value reported by CGAL in \cref{tab:full_results}. Across the 41 instances, this ratio averages $0.87$ and has median $0.94$. As a rough comparison, the scale $1/(c_0^2\log d)$ suggested by the SSR-based guarantee (ignoring the hidden absolute constant) averages $0.13$, and the empirical ratio exceeds this scale on every instance. Thus, the SSR-based prediction is conservative, but it still provides useful explanatory power for the numerical behavior.

    \paragraph{Detailed results.}
    Comprehensive comparisons among the methods are detailed in \cref{tab:full_results}.
     We use ``Obj'' to denote the objective value of a solution found by a certain algorithm, ``Time'' to denote the runtime of an algorithm.
     Note that we run \cref{alg:multi_ra} with $N = 3000$, and report the largest objective value among those feasible solutions found in the column ``Obj'' belonging to ``\cref{alg:multi_ra}''. 
     For reproducibility, we set the random seed to 42.
    In the table, $c_0$ stands for $\textup{SSR} / \sqrt{k}$, measuring how far $\textup{SSR}$ is from $\sqrt{k}$.
    The column ``Total Time'' stands for the runtime of CGAL + \cref{alg:multi_ra}.
    We highlight the objective values that are largest among all algorithms.

    \begin{landscape}            
  \begin{table}[p]
    \caption{Complete numerical results for all 41 instances. Objective values and runtimes for each algorithm are reported.
    The time limit for Branch-and-Bound algorithm is set to 700 seconds.
    In the columns belonging to ``CGAL'', objective values for \ref{prob SDP SPCA} are reported, and $c_0$ stands for $\textup{SSR} / \sqrt{k}$. In the columns belonging to ``Algorithm~\ref{alg:multi_ra}'', ``Time'' stands for the runtime of \cref{alg:multi_ra} only, while ``Total Time'' stands for the runtime of CGAL + \cref{alg:multi_ra}.}
    \label{tab:full_results}
    \small               
    \centering
    \adjustbox{max width=1.3\textwidth}{%
      \begin{tabular}{lccccccccccccccccccccc}
\toprule
Dataset & d & k & \multicolumn{2}{c}{Greedy} & \multicolumn{2}{c}{Local Search} & \multicolumn{2}{c}{Chan's} & \multicolumn{2}{c}{Low Rank} & \multicolumn{2}{c}{Branch-and-Bound} & \multicolumn{3}{c}{MSPCA} & \multicolumn{3}{c}{CGAL} & \multicolumn{3}{c}{\cref{alg:multi_ra}} \\ 
\cmidrule(lr){4-5}
\cmidrule(lr){6-7}
\cmidrule(lr){8-9}
\cmidrule(lr){10-11}
\cmidrule(lr){12-13}
\cmidrule(lr){14-16}
\cmidrule(lr){17-19}
\cmidrule(lr){20-22}
 &  &  & Obj & Time & Obj & Time & Obj & Time & Obj & Time & Obj & Time & LB & UB & Time & Time & SDP Obj & $c_0$ & Obj & Time & Total Time \\ 
\midrule
Eisen1 & 79 & 2 & \textbf{7.583} & 0.005 & \textbf{7.583} & 0.013 & 7.582 & 0.003 & \textbf{7.583} & 0.304 & \textbf{7.583} & 0.015 & \textbf{7.583} & 7.825 & 5.675 & 0.077 & 7.820 & 1.320 & \textbf{7.583} & 0.296 & 0.372 \\
Eisen1 & 79 & 5 & \textbf{14.076} & 0.015 & \textbf{14.076} & 0.032 & 14.072 & 0.003 & \textbf{14.076} & 0.343 & \textbf{14.076} & 0.078 & \textbf{14.076} & 14.272 & 11.131 & 0.071 & 14.623 & 1.023 & \textbf{14.076} & 0.349 & 0.420 \\
Eisen1 & 79 & 10 & \textbf{17.335} & 0.031 & \textbf{17.335} & 0.063 & \textbf{17.335} & 0.003 & \textbf{17.335} & 0.399 & \textbf{17.335} & 0.508 & 16.415 & 17.399 & 27.938 & 0.073 & 17.712 & 1.006 & \textbf{17.335} & 0.380 & 0.453 \\
Eisen1 & 79 & 20 & \textbf{17.719} & 0.058 & \textbf{17.719} & 0.127 & \textbf{17.719} & 0.003 & \textbf{17.719} & 0.503 & \textbf{17.719} & 700.000 & 14.530 & 17.799 & 117.359 & 0.046 & 18.131 & 0.967 & \textbf{17.719} & 0.417 & 0.463 \\
Eisen1 & 79 & 50 & \textbf{18.069} & 0.140 & \textbf{18.069} & 0.241 & \textbf{18.069} & 0.004 & \textbf{18.069} & 0.760 & \textbf{18.069} & 700.001 & 18.030 & 18.082 & 121.533 & 0.044 & 18.131 & 0.612 & \textbf{18.069} & 0.530 & 0.574 \\
News & 100 & 2 & 2750.539 & 0.008 & 2750.539 & 0.021 & 2749.795 & 0.010 & \textbf{2750.540} & 0.282 & \textbf{2750.540} & 0.005 & \textbf{2750.540} & 2751.281 & 12.496 & 0.091 & 2745.233 & 1.521 & \textbf{2750.540} & 0.419 & 0.510 \\
News & 100 & 5 & \textbf{3557.755} & 0.030 & 3557.754 & 0.069 & 3483.899 & 0.010 & 2861.836 & 0.300 & \textbf{3557.755} & 1.531 & \textbf{3557.755} & 3673.218 & 70.845 & 0.094 & 3679.519 & 1.126 & \textbf{3557.755} & 0.543 & 0.637 \\
News & 100 & 10 & 4549.526 & 0.061 & 4549.526 & 0.123 & 4523.123 & 0.009 & 3171.891 & 0.316 & \textbf{4549.527} & 700.000 & \textbf{4549.527} & 4727.807 & 114.517 & 0.090 & 4756.311 & 1.043 & \textbf{4549.527} & 0.511 & 0.601 \\
News & 100 & 20 & 5696.468 & 0.116 & \textbf{5696.469} & 0.229 & 5689.103 & 0.009 & 3674.691 & 0.336 & \textbf{5696.469} & 700.000 & \textbf{5696.469} & 5935.193 & 125.960 & 0.085 & 6050.354 & 1.011 & \textbf{5696.469} & 0.517 & 0.602 \\
News & 100 & 50 & 7017.849 & 0.266 & 7017.850 & 0.457 & 7015.768 & 0.010 & 4126.741 & 0.290 & \textbf{7017.851} & 700.000 & \textbf{7017.851} & 7147.107 & 148.708 & 0.095 & 7334.629 & 1.000 & 7017.103 & 0.587 & 0.682 \\
News & 100 & 100 & \textbf{7367.672} & 0.370 & \textbf{7367.672} & 0.373 & \textbf{7367.672} & 0.009 & \textbf{7367.672} & 0.316 & 7367.667 & 0.000 & \textbf{7367.672} & 7367.672 & 0.003 & 0.049 & 7367.672 & 0.755 & \textbf{7367.672} & 0.749 & 0.799 \\
Eisen2 & 118 & 2 & \textbf{3.976} & 0.009 & \textbf{3.976} & 0.020 & \textbf{3.976} & 0.013 & \textbf{3.976} & 0.655 & \textbf{3.976} & 0.034 & \textbf{3.976} & 3.989 & 10.164 & 0.107 & 4.167 & 2.103 & \textbf{3.976} & 0.320 & 0.427 \\
Eisen2 & 118 & 5 & 6.425 & 0.023 & 6.425 & 0.054 & 6.635 & 0.013 & \textbf{6.636} & 0.763 & \textbf{6.636} & 2.622 & \textbf{6.636} & 6.824 & 20.064 & 0.109 & 7.117 & 1.892 & 6.541 & 0.404 & 0.512 \\
Eisen2 & 118 & 10 & 11.412 & 0.048 & \textbf{11.718} & 0.196 & 11.715 & 0.014 & \textbf{11.718} & 0.835 & \textbf{11.718} & 315.274 & 11.441 & 11.844 & 26.214 & 0.108 & 11.970 & 1.432 & \textbf{11.718} & 0.385 & 0.492 \\
Eisen2 & 118 & 20 & \textbf{19.323} & 0.100 & \textbf{19.323} & 0.220 & 19.312 & 0.014 & \textbf{19.323} & 1.058 & \textbf{19.323} & 700.000 & \textbf{19.323} & 19.597 & 38.332 & 0.106 & 19.559 & 1.155 & \textbf{19.323} & 0.439 & 0.544 \\
Eisen2 & 118 & 50 & \textbf{26.035} & 0.257 & \textbf{26.035} & 0.515 & 26.025 & 0.014 & 26.031 & 1.883 & \textbf{26.035} & 700.000 & 25.831 & 26.280 & 145.161 & 0.109 & 27.083 & 1.003 & \textbf{26.035} & 0.575 & 0.684 \\
Eisen2 & 118 & 100 & \textbf{27.573} & 0.451 & \textbf{27.573} & 0.674 & \textbf{27.573} & 0.014 & \textbf{27.573} & 2.250 & \textbf{27.573} & 700.000 & 27.471 & 27.597 & 172.536 & 0.060 & 27.679 & 0.821 & \textbf{27.573} & 0.749 & 0.810 \\
CovColon & 500 & 2 & \textbf{715.395} & 0.038 & \textbf{715.395} & 0.113 & \textbf{715.395} & 0.160 & \textbf{715.395} & 13.481 & \textbf{715.395} & 0.040 & \textbf{715.395} & 817.226 & 33.006 & 1.955 & 4628.146 & 13.003 & \textbf{715.395} & 0.553 & 2.508 \\
CovColon & 500 & 5 & \textbf{1646.454} & 0.096 & \textbf{1646.454} & 0.277 & 1646.412 & 0.163 & \textbf{1646.454} & 14.172 & \textbf{1646.454} & 0.172 & \textbf{1646.454} & 1664.632 & 78.784 & 2.132 & 5110.411 & 7.487 & \textbf{1646.454} & 0.639 & 2.771 \\
CovColon & 500 & 10 & \textbf{2641.229} & 0.181 & \textbf{2641.229} & 0.557 & 2640.302 & 0.166 & \textbf{2641.229} & 17.591 & \textbf{2641.229} & 700.000 & 2627.404 & 2735.330 & 156.236 & 1.878 & 5819.470 & 5.374 & \textbf{2641.229} & 0.667 & 2.545 \\
CovColon & 500 & 20 & 4255.287 & 0.398 & \textbf{4255.694} & 1.962 & 4253.598 & 0.166 & \textbf{4255.694} & 26.237 & \textbf{4255.694} & 700.000 & 4250.320 & 4420.541 & 149.264 & 2.052 & 7291.615 & 3.502 & 4254.765 & 0.693 & 2.745 \\
CovColon & 500 & 50 & \textbf{7977.493} & 1.138 & \textbf{7977.493} & 3.342 & 7977.381 & 0.166 & \textbf{7977.493} & 65.019 & \textbf{7977.493} & 700.000 & 7973.697 & 8232.498 & 273.509 & 1.881 & 12175.880 & 1.889 & 7976.408 & 0.803 & 2.683 \\
CovColon & 500 & 100 & \textbf{12307.385} & 3.010 & \textbf{12307.385} & 8.528 & 12307.080 & 0.163 & 12307.228 & 208.050 & \textbf{12307.385} & 700.003 & 12256.380 & 12511.080 & 446.110 & 1.957 & 16957.169 & 1.437 & \textbf{12307.385} & 1.125 & 3.081 \\
LymphomaCov & 500 & 2 & \textbf{2064.868} & 0.036 & \textbf{2064.868} & 0.119 & 2064.864 & 0.184 & \textbf{2064.868} & 15.086 & \textbf{2064.868} & 0.233 & \textbf{2064.868} & 2652.215 & 173.575 & 2.119 & 3223.160 & 10.030 & \textbf{2064.868} & 0.625 & 2.744 \\
LymphomaCov & 500 & 5 & 3782.621 & 0.094 & 3782.621 & 0.279 & 4297.340 & 0.180 & \textbf{4300.497} & 16.836 & \textbf{4300.497} & 40.169 & \textbf{4300.497} & 4375.991 & 101.530 & 1.831 & 5080.984 & 4.933 & \textbf{4300.497} & 0.734 & 2.565 \\
LymphomaCov & 500 & 10 & 5911.412 & 0.189 & 5911.412 & 0.556 & 5909.077 & 0.191 & \textbf{6008.741} & 19.093 & \textbf{6008.741} & 700.000 & \textbf{6008.741} & 6857.551 & 163.512 & 1.797 & 6893.303 & 3.311 & 6008.317 & 0.769 & 2.566 \\
LymphomaCov & 500 & 20 & 9063.961 & 0.439 & \textbf{9082.158} & 2.449 & 8979.248 & 0.188 & 9063.961 & 24.743 & \textbf{9082.158} & 700.001 & \textbf{9082.158} & 9916.811 & 153.780 & 1.811 & 10077.055 & 2.276 & 9063.960 & 0.770 & 2.580 \\
LymphomaCov & 500 & 50 & 14546.930 & 1.471 & \textbf{14546.931} & 3.997 & 14388.111 & 0.186 & 14541.948 & 64.599 & \textbf{14546.931} & 700.002 & 13763.608 & 15338.658 & 241.648 & 2.139 & 15541.962 & 1.389 & \textbf{14546.931} & 0.916 & 3.055 \\
LymphomaCov & 500 & 100 & \textbf{19339.870} & 3.696 & \textbf{19339.870} & 9.875 & 19162.191 & 0.178 & 19253.991 & 210.344 & \textbf{19339.870} & 700.000 & 16668.065 & 19260.237 & 370.315 & 1.983 & 20941.087 & 1.129 & 19336.097 & 1.188 & 3.171 \\
Reddit1500 & 1500 & 2 & \textbf{920.074} & 0.091 & \textbf{920.074} & 0.543 & 920.044 & 1.552 & \textbf{920.074} & 118.997 & \textbf{920.074} & 7.457 & \textbf{920.074} & 942.504 & 84.123 & 2.474 & 978.858 & 1.201 & \textbf{920.074} & 2.951 & 5.424 \\
Reddit1500 & 1500 & 5 & \textbf{980.974} & 0.269 & \textbf{980.974} & 1.499 & 979.129 & 1.478 & \textbf{980.974} & 125.956 & \textbf{980.974} & 700.010 & \textbf{980.974} & 1018.085 & 133.564 & 2.177 & 1080.998 & 1.105 & \textbf{980.974} & 3.740 & 5.917 \\
Reddit1500 & 1500 & 10 & \textbf{1045.743} & 0.631 & \textbf{1045.743} & 3.056 & 1042.357 & 1.534 & 1044.777 & 126.630 & \textbf{1045.743} & 700.002 & \textbf{1045.743} & 1077.079 & 167.336 & 2.347 & 1146.300 & 1.061 & \textbf{1045.743} & 3.310 & 5.657 \\
Reddit1500 & 1500 & 20 & \textbf{1105.286} & 1.463 & \textbf{1105.286} & 6.859 & 1103.282 & 1.552 & 1091.657 & 136.850 & \textbf{1105.286} & 700.007 & 1096.453 & 1132.024 & 231.820 & 5.545 & 1193.533 & 1.024 & 1105.068 & 5.627 & 11.172 \\
Reddit1500 & 1500 & 50 & \textbf{1172.655} & 5.057 & \textbf{1172.655} & 24.446 & 1171.323 & 1.529 & 1130.934 & 181.040 & \textbf{1172.655} & 700.012 & 1151.347 & 1171.751 & 280.928 & 2.546 & 1230.595 & 1.007 & \textbf{1172.655} & 4.187 & 6.734 \\
Reddit1500 & 1500 & 100 & 1200.142 & 13.617 & \textbf{1200.751} & 87.973 & 1200.435 & 1.586 & 1140.451 & 501.478 & \textbf{1200.751} & 700.012 & 1165.997 & 1187.112 & 297.662 & 2.345 & 1241.312 & 1.000 & 1200.663 & 5.527 & 7.871 \\
Reddit2000 & 2000 & 2 & \textbf{1254.755} & 0.121 & \textbf{1254.755} & 0.863 & 1253.590 & 3.643 & \textbf{1254.755} & 257.563 & \textbf{1254.755} & 17.952 & \textbf{1254.755} & 1353.007 & 118.676 & 2.568 & 1352.753 & 1.526 & \textbf{1254.755} & 5.671 & 8.239 \\
Reddit2000 & 2000 & 5 & \textbf{1397.358} & 0.463 & \textbf{1397.358} & 2.511 & 1369.780 & 3.753 & \textbf{1397.358} & 261.937 & \textbf{1397.358} & 700.008 & \textbf{1397.358} & 1543.256 & 134.227 & 2.903 & 1545.370 & 1.238 & \textbf{1397.358} & 6.195 & 9.099 \\
Reddit2000 & 2000 & 10 & 1521.308 & 1.073 & 1521.308 & 5.237 & 1482.320 & 3.705 & 1504.450 & 265.079 & \textbf{1523.823} & 700.012 & \textbf{1523.823} & 1731.351 & 130.650 & 2.848 & 1733.370 & 1.130 & \textbf{1523.823} & 7.130 & 9.978 \\
Reddit2000 & 2000 & 20 & 1670.471 & 2.738 & 1684.394 & 44.551 & 1666.240 & 3.617 & 1612.458 & 271.172 & \textbf{1684.395} & 700.012 & 1670.153 & 1961.127 & 164.186 & 2.723 & 2026.850 & 1.076 & 1684.394 & 6.459 & 9.182 \\
Reddit2000 & 2000 & 50 & 2289.037 & 7.926 & 2322.820 & 94.621 & 2311.241 & 3.751 & 1967.200 & 331.002 & \textbf{2322.821} & 700.007 & 2218.043 & 2315.127 & 288.524 & 2.831 & 2502.577 & 1.016 & 2322.820 & 6.892 & 9.723 \\
Reddit2000 & 2000 & 100 & 2544.370 & 19.426 & 2544.370 & 72.566 & 2536.516 & 3.685 & 1841.806 & 654.783 & \textbf{2544.371} & 700.007 & 2269.553 & 2338.926 & 349.681 & 2.898 & 2600.108 & 1.002 & 2543.924 & 9.057 & 11.955 \\
\midrule 
Average &  &  & 3230.839 & 1.601 & 3232.479 & 9.273 & 3228.698 & 0.820 & 3043.605 & 96.081 & 3247.551 & 453.322 & 3150.170 & 3373.851 & 143.690 & 1.443 & 4011.721 & 2.169 & 3246.916 & 2.058 & 3.501 \\
\bottomrule
\end{tabular}
   
    }
  \end{table}
\end{landscape}

\ifthenelse{\boolean{IJOO}}{}{\section{Acknowledgements}
A. Del Pia and D. Zhou are partially funded by AFOSR grant FA9550-23-1-0433. 
Any opinions, findings, and conclusions or recommendations expressed in this material are those of the authors and do not necessarily reflect the views of the Air Force Office of Scientific Research.
The authors would love to thank Agniva Chowdhury and Lijun Ding for the discussions on algorithms to solve \ref{prob SDP SPCA}.}
\IJOOPrintBibliography
\ifIJOOincludeAppendix
\clearpage
\IJOOInputProofAppendix
\fi
\else
\ifIJOOincludeAppendix
\IJOOInputProofAppendix
\clearpage
\fi
\IJOOPrintBibliography
\fi


\end{document}